\documentclass[lettersize,journal]{IEEEtran}

\usepackage{graphicx}
\usepackage[edges]{forest}
\usepackage{tikz-qtree}

\usepackage{amsmath,amsfonts}
\usepackage{algorithmic}
\usepackage{algorithm}
\usepackage{array}
\usepackage[caption=false,font=normalsize,labelfont=sf,textfont=sf]{subfig}
\usepackage{textcomp}
\usepackage{stfloats}
\usepackage{url}
\usepackage{verbatim}
\usepackage{graphicx}
\usepackage{cite}
\usepackage{forest}

\usepackage{booktabs}
\usepackage{multirow}
\usepackage{multicol}
\usepackage{tabularx}
\usepackage{float}
\usepackage{makecell}
\usepackage{enumitem}
\usepackage{bm}
\usepackage{graphicx}
\usepackage{color}
\usepackage{soul}
\usepackage{caption}
\usepackage{subcaption}

\def\lightblue#1{\textcolor[rgb]{0.3,0.3,1}{#1}}

\definecolor{hidden-draw}{rgb}{0.5, 0.5, 0.5}
\definecolor{harvestgold}{rgb}{0.85, 0.57, 0.0}
\definecolor{cyan}{rgb}{0.0, 1.0, 1.0}
\definecolor{lightcoral}{rgb}{0.94, 0.5, 0.5}
\definecolor{SlateBlue}{rgb}{0.416, 0.353, 0.804}
\definecolor{LightGreen}{rgb}{0.564, 0.933, 0.564}
\definecolor{Turquoise}{rgb}{0.251, 0.878, 0.816}
\definecolor{BlueGreen}{rgb}{0.643, 0.859, 0.871}
\definecolor{BlueViolet}{rgb}{0.54, 0.17, 0.89}
\definecolor{RAG}{rgb}{0.639, 0.835, 0}
\definecolor{0}{rgb}{0.92, 0.3, 0.26}
\definecolor{1}{rgb}{0.961, 0.851, 0.471}
\definecolor{1_1}{rgb}{1.00, 0.933, 0.698}
\definecolor{4}{rgb}{0.522, 0.784, 0.953}
\definecolor{4_1}{rgb}{0.722, 0.953, 0.953}
\definecolor{3}{rgb}{0.56, 0.93, 0.56}
\definecolor{3_1}{rgb}{0.808, 0.996, 0.808}
\definecolor{2}{rgb}{0.988, 0.541, 0.565}
\definecolor{2_1}{rgb}{0.996, 0.796, 0.808}
\definecolor{bounding}{rgb}{0.855, 0.392, 0.357}
\definecolor{5}{rgb}{1.0, 0.8, 0.6}
\definecolor{6}{rgb}{0.8, 0.6, 1.0}
\definecolor{7}{rgb}{0.98, 0.68, 0.68} 
\definecolor{8}{rgb}{0.98, 0.87, 0.71}
\definecolor{9}{rgb}{0.72, 0.91, 0.85} 
\definecolor{10}{rgb}{0.20, 0.40, 0.80}   
\definecolor{11}{rgb}{0.98, 0.85, 0.42}   
\definecolor{12}{rgb}{0.99, 0.76, 0.75}

\usepackage{booktabs}
\usepackage{pifont}
\usepackage{xcolor} 
\usepackage{url}
\usepackage{tikz}

\usetikzlibrary{backgrounds,mindmap}
\usetikzlibrary{calc,positioning,intersections}
\usepackage{pgfplots}
\usepackage{listings}
\usepackage{multirow}
\usepackage{amsmath}

\definecolor{lightcoral}{rgb}{0.94, 0.5, 0.5}
\definecolor{lightgreen}{rgb}{0.56, 0.93, 0.56}
\definecolor{harvestgold}{rgb}{0.98, 0.85, 0.40}
\definecolor{brightlavender}{rgb}{0.75, 0.58, 0.89}
\definecolor{capri}{rgb}{0.0, 0.75, 1.0}
\definecolor{carminepink}{rgb}{0.92, 0.3, 0.26}
\definecolor{celadon}{rgb}{0.67, 0.88, 0.69}
\definecolor{darkpastelgreen}{rgb}{0.01, 0.75, 0.24}

\definecolor{hidden-draw}{RGB}{205, 44, 36}
\definecolor{hidden-blue}{RGB}{194,232,247}
\definecolor{hidden-orange}{RGB}{243,202,120}
\definecolor{hidden-yellow}{RGB}{242,244,193}
\definecolor{tree-level-1}{RGB}{245,20,85}
\definecolor{tree-level-2}{RGB}{246,86,118}
\definecolor{tree-level-3}{RGB}{248,177,193}
\definecolor{tree-leaf}{RGB}{176,230,198}

\definecolor{Self}{RGB}{255,0,128}
\definecolor{Ensemble}{RGB}{0,127,255}
\definecolor{Iterative}{RGB}{153,51,255}

\definecolor{exemplar1}{RGB}{136,98,148}
\definecolor{exemplar2}{RGB}{148,210,242}
\definecolor{knowledge1}{RGB}{249,219,152}
\definecolor{knowledge2}{RGB}{255,245,220}
\usepackage{hyperref} 
\tikzstyle{my-box}=[
    rectangle,
    draw=hidden-draw,
    rounded corners,
    text opacity=1,
    minimum height=1.5em,
    minimum width=5em,
    inner sep=2pt,
    align=center,
    fill opacity=.5,
]
\tikzstyle{cause_leaf}=[my-box, minimum height=1.5em,
    fill=harvestgold!20, text=black, align=left,font=\scriptsize,
    inner xsep=2pt,
    inner ysep=4pt,
]
\tikzstyle{detect_leaf}=[my-box, minimum height=1.5em,
    fill=cyan!20, text=black, align=left,font=\scriptsize,
    inner xsep=2pt,
    inner ysep=4pt,
]
\tikzstyle{mitigate_leaf}=[my-box, minimum height=1.5em,
    fill=lightcoral!20, text=black, align=left,font=\scriptsize,
    inner xsep=2pt,
    inner ysep=4pt,
]
\tikzstyle{mitigate_leaf}=[my-box, minimum height=1.5em,
    fill=lightcoral!20, text=black, align=left,font=\scriptsize,
    inner xsep=2pt,
    inner ysep=4pt,
]
\tikzstyle{llm_leaf}=[my-box, minimum height=1.5em,
    fill=lightgreen!20, text=black, align=left,font=\scriptsize,
    inner xsep=2pt,
    inner ysep=4pt,
]
\tikzstyle{challenge_leaf}=[my-box, minimum height=1.5em,
    fill=brightlavender!20, text=black, align=left,font=\scriptsize,
    inner xsep=2pt,
    inner ysep=4pt,
]

\begin{document}

\title{A Survey on Remote Sensing Foundation Models: From Vision to Multimodality}

\author{Ziyue~Huang, Hongxi~Yan, Qiqi~Zhan, Shuai~Yang, Mingming~Zhang, Chenkai~Zhang, YiMing~Lei, Zeming~Liu, Qingjie~Liu, ~\IEEEmembership{Member,~IEEE}, and Yunhong~Wang,~\IEEEmembership{Fellow,~IEEE}%

\thanks{This work was supported by the National Natural Science Foundation of China	under Grant 62176017. \textit{(Corresponding author: Qingjie Liu)}}
\thanks{
Ziyue~Huang, Hongxi~Yan, Qiqi~Zhan, Shuai~Yang, Mingming~Zhang, Chenkai~Zhang, YiMing~Lei, Zeming~Liu, Qingjie~Liu, and Yunhong Wang are with the State Key Laboratory of Virtual Reality Technology and Systems, Beihang University, Beijing 100191, China, and also with the Hangzhou Innovation Institute, Beihang University, Hangzhou 310051, China.}
}




\maketitle
\begin{abstract}
The rapid advancement of remote sensing foundation models, particularly vision and multimodal models, has significantly enhanced the capabilities of intelligent geospatial data interpretation. 
These models combine various data modalities, such as optical, radar, and LiDAR imagery, with textual and geographic information, enabling more comprehensive analysis and understanding of remote sensing data. 
The integration of multiple modalities allows for improved performance in tasks like object detection, land cover classification, and change detection, which are often challenged by the complex and heterogeneous nature of remote sensing data. 
However, despite these advancements, several challenges remain. The diversity in data types, the need for large-scale annotated datasets, and the complexity of multimodal fusion techniques pose significant obstacles to the effective deployment of these models. Moreover, the computational demands of training and fine-tuning multimodal models require significant resources, further complicating their practical application in remote sensing image interpretation tasks. 
This paper provides a comprehensive review of the state-of-the-art in vision and multimodal foundation models for remote sensing, focusing on their architecture, training methods, datasets and application scenarios. We discuss the key challenges these models face, such as data alignment, cross-modal transfer learning, and scalability, while also identifying emerging research directions aimed at overcoming these limitations. 
Our goal is to provide a clear understanding of the current landscape of remote sensing foundation models and inspire future research that can push the boundaries of what these models can achieve in real-world applications. The list of resources collected by the paper can be found in the \href{https://github.com/IRIP-BUAA/A-Review-for-remote-sensing-vision-language-models}{https://github.com/IRIP-BUAA/A-Review-for-remote-sensing-vision-language-models}. 
\end{abstract}


\section{Introduction}
In recent years, significant advances in deep learning and artificial intelligence have solidified these technologies as critical tools for intelligent interpretation in the field of remote sensing. 
Deep learning has been widely applied to a range of geospatial tasks, such as scene classification \cite{sydney, zhao2015dirichlet, cheng2017remote, xia2017aid, rs-clip}, object detection \cite{diorr, dior, dota, SODA, zhang2021sar}, change detection \cite{chen2020spatial, daudt2018urban, change-agent}, land cover classification \cite{helber2019eurosat, loveda, Scheibenreif_2022_CVPR, yang2010bag}, and geospatial localization \cite{satclip, geoclip, xu2024addressclip}. 
However, most current models are specifically designed for particular tasks, with highly task-oriented architecture \cite{cheng2024change}, loss functions \cite{yang2021rethinking}, and training strategies \cite{wang2020weakly}. 
This high degree of specialization substantially constrains the generalization of these models, even across closely related tasks \cite{zhou2024towards}. 
Furthermore, these models underutilize the vast volumes of available remote sensing data, leading to suboptimal generalization and diminished performance in practical applications.

The rise of foundation models, driven by the rapid advances in self-supervised and multimodal learning, has fundamentally reshaped the landscape of artificial intelligence.
In this article, foundation models refer to deep learning models that are pretrained on extensive datasets using techniques such as self-supervised, semi-supervised, or multimodal learning to extract general features, and can be efficiently adapted to various downstream tasks through fine-tuning or prompt-tuning.
Leveraging vast parameters and comprehensive data, these models have not only achieved revolutionary progress in natural language processing (e.g., the GPT series, LLaMA, and other large language models) but have also shown remarkable generalization in computer vision and other fields, thereby propelling the applications of artificial intelligence.

Foundation models were first popularized in natural language processing (NLP) through pivotal breakthroughs such as GPT \cite{gpt, gpt2, gpt3}, LLaMA \cite{touvron2023llama2}, and other large language models (LLMs).
These models employ multi-stage training with billions of parameters and vast amounts of textual data, achieving state-of-the-art performance in tasks such as language understanding, text generation, and machine translation, and demonstrating generalization in zero-shot and few-shot learning \cite{zhao2023survey}. 
In vision tasks, foundational models such as DINOv2 \cite{dinov2} achieve effective zero-shot image retrieval by leveraging self-supervised learning, trained on diverse web-scale datasets. 
SAM \cite{sam} employs a semi-supervised training pipeline to develop a highly reliable foundation model for prompt-based segmentation. 
However, due to the complexity and diversity of vision tasks, these models still require additional fine-tuning or task-specific modules to adjust to downstream tasks \cite{lin2024clip, chen2024rsprompter}. 
Vision-language models (VLMs), such as CLIP \cite{clip} and Grounding DINO \cite{liu2025grounding}, achieve extensive alignment between text and image data, enabling the models to perform zero-shot inference through textual prompts. 
Multimodal Large Language Models (MLLMs), such as GPT-4V \cite{achiam2023gpt}, convert both images and text into a unified token sequence for consistent processing, enabling more flexible handling of various downstream tasks. 

Remote sensing data presents unique challenges for intelligent interpretation. 
The scarcity of labeled data and the specialized requirements of remote sensing tasks have historically constrained the performance of deep learning models in this domain \cite{remoteclip, geoclip}. 
Image interpretation is a core task in remote sensing, making vision and multimodal foundational models the most prevalent. 
Existing research typically adapts general models by incorporating task-specific data, structures, and training strategies relevant to remote sensing interpretation. 
Vision foundation models build upon conventional Convolutional Neural Networks (CNNs) and Transformer architectures, incorporating enhancements tailored to remote sensing scenarios. 
By designing specialized pretraining tasks and model architectures, these models can more effectively extract visual representations from high-resolution images \cite{diao2024ringmo}.
Multimodal foundation models integrate visual and textual data, offering new solutions for remote sensing tasks. 
VLM models based on CLIP map remote sensing images to textual descriptions, enabling scene classification and retrieval through text prompts \cite{remoteclip}.
Meanwhile, multimodal models leveraging MLLM architectures support a broader range of tasks, demonstrating enhanced generalizability.

To provide a thorough understanding of the current state of remote sensing foundation models, this survey systematically reviews recent advancements, identifies key challenges, and outlines future research directions.
Unlike previous surveys that focus on specific tasks or stages, this study adopts a comprehensive perspective, examining the architecture, training methodologies, data utilization, and application scenarios of foundation models.
The key contributions of this survey are as follows:
\begin{itemize}
    \item Comprehensive Survey: This work presents the first comprehensive survey dedicated to remote sensing foundation models, covering both vision and multimodal approaches. It systematically reviews their evolution, technical innovations, and key achievements.
    \item Innovative Taxonomy: We introduce a novel organizational framework that classifies research from two perspectives: model architecture and primary functionality. This taxonomy provides a structured understanding of the development, interconnections, and applicability of various approaches.
    \item Resource Compilation: To support ongoing research, we compile and maintain a dedicated repository of resources, including curated papers, leaderboards, and open-source code, fostering collaboration and innovation in the field.
\end{itemize}

\begin{figure*}[t]
    \centering
    \resizebox{\textwidth}{!}{
    \begin{forest}
        forked edges, 
        for tree={
            grow=east, 
            reversed=true, 
            anchor=base west, 
            parent anchor=east, 
            child anchor=west, 
            base=left, 
            font=\small, 
            rectangle, 
            draw=hidden-draw, 
            rounded corners, 
            align=left, 
            minimum width=4em, 
            edge+={darkgray,  line width=1pt}, 
            s sep=3pt, 
            inner xsep=2pt, 
            inner ysep=3pt, 
            ver/.style={rotate=90,  child anchor=north,  parent anchor=south,  anchor=center}, 
        }, 
        where level=1{text width=6.3em, font=\scriptsize, }{}, 
        where level=2{text width=7.3em, font=\scriptsize, }{}, 
        where level=3{text width=9.5em, font=\scriptsize, }{}, 
        where level=4{text width=8.0em, font=\scriptsize, }{}, 
        [Remote Sensing Foundation Model,  ver,  color=carminepink!100,  fill=carminepink!15,  text=black
            [Architecture,  color=5!100,  fill=5!100,  text=black
                [Vision Models,  color=1!100,  fill=1_1!100,  text=black
                    [CNN-based,  color=1!100,  fill=1_1!100,  text=black
                        [{\,  SMLFR \cite{10378718} and MMEarth \cite{nedungadi2024mmearth}},  cause_leaf,  text width=34.0em
                        ]
                    ]
                    [Transformer-Based,  color=1!100,  fill=1_1!100,  text=black
                        [ {\,  SkySense \cite{guo2024skysense},  Prithvi \cite{jakubik2023foundationmodelsgeneralistgeospatial},  Presto \cite{tseng2024lightweight},  EarthPT \cite{smith2024earthpt},  USat \cite{irvin2023usat},  OFA-Net \cite{xiong2024all},  DOFA \cite{xiong2024neural},   \\ \,  RSPrompter \cite{chen2023rsprompter},  LeMeViT \cite{jiang2024lemevit} ,  HyperSIGMA \cite{wang2024hypersigma}},  cause_leaf,  text width=34.0em
                        ]
                    ]
                    [CNN-Transformer Hybrid,  color=1!100,  fill=1_1!100,  text=black
                        [{\,  U-BARN \cite{10414422},  RingMo-lite \cite{wang2023ringmolite}},  cause_leaf,  text width=34.0em
                        ]
                    ]
                ]
                [Multimodal Models,  color=2!100,  fill=2_1!100,  text=black
                    [CLIP-based Models,  color=2!100,  fill=2_1!100,  text=black
                        [{\,   RemoteCLIP\cite{remoteclip},  Skyscript\cite{skyscript},  GeoRSCLIP\cite{georsclip},  S-CLIP\cite{s-clip},  RS-CLIP\cite{rs-clip},  Mall et al.\cite{remoteWoAnn},  \\ \, Yang et al.\cite{bootstrapping},  GeoCLIP\cite{geoclip},  SatCLIP\cite{satclip},  AddressCLIP\cite{xu2024addressclip},  StreetCLIP\cite{haas2023learning},  GeoCLAP\cite{geoclap}},  cause_leaf,  text width=34.0em
                        ]
                    ]
                    [Diffusion-based Models,  color=2!100,  fill=2_1!100,  text=black
                        [{\,  RSDiff\cite{sebaq2023rsdiff},  DiffusionSat\cite{diffusionsat},  CRS-Diff\cite{Crs-diff},  MetaEarth\cite{metaearth}},  cause_leaf,  text width=34.0em
                        ]
                    ]
                    [MLLM,  color=2!100,  fill=2_1!100,  text=black
                        [{\,  LHRS-Bot\cite{lhrs-bot},  H2RSVLM\cite{h2rsvlm},  Geochat\cite{geochat}, Skysensegpt\cite{skysensegpt},  RSGPT\cite{rsgpt},  SkyEyeGPT\cite{skyeyegpt},  \\ \, EarthGPT\cite{earthgpt},   Popeye\cite{popeye},  RS-CapRet\cite{rs-cap},  RSGPT\cite{rsgpt},  RS-LLaVA\cite{rs-llava}},  cause_leaf,  text width=34.0em
                        ]
                    ]
                    [LLM-based Agent,  color=2!100,  fill=2_1!100,  text=black
                        [{\,  Tree-GPT\cite{treegpt},  Remote Sensing ChatGPT\cite{remote-sense-chatgpt},  Change-Agent\cite{change-agent},  RS-Agent\cite{rs-agent}},  cause_leaf,  text width=34.0em
                        ]
                    ]
                ]
            ]
            [Training Method,  color=6!100,  fill=6!100,  text=black
                [Vision Models,  color=3!100,  fill=3_1!100,  text=black
                    [Contrastive Method,  color=3!100,  fill=3_1!100,  text=black
                        [{\,  CMC-RSSR \cite{Stojnic_2021_CVPR},  DINO-MC \cite{Wanyan_2024_CVPR},  CACo\cite{Mall_2023_CVPR}, GASSL \cite{Ayush_2021_ICCV},  SeCo \cite{Manas_2021_ICCV}, MATTER \cite{Akiva_2022_CVPR}, \\ \, DINO-MC \cite{Wanyan_2024_CVPR},   SSLTransformerRS \cite{Scheibenreif_2022_CVPR},  DINO-MM \cite{wang2022selfsupervised},  RSBYOL \cite{rs_byol_2022},  IaI-SimCLR \cite{Prexl_2023_CVPR},\\ \,  DeCUR \cite{wang2023decurdecouplingcommon},  SkySense \cite{han2024bridging}},  cause_leaf,  text width=34.0em
                        ]
                    ]
                    [Generative Method,  color=3!100,  fill=3_1!100,  text=black
                        [ {\,  Scale-MAE \cite{10377166},  SatMAE++ \cite{noman2024rethinking},  SMLFR \cite{10453587},  SAR-JEPA \cite{li2024predictinggradientbetterexploring}, SatMAE \cite{satmae2022},  RingMo \cite{ringmo_2023},  \\ \, RingMo-Sense \cite{10254320},  CtxMIM \cite{zhang2024ctxmimcontextenhancedmaskedimage},  S2MAE \cite{Li_2024_CVPR},  SpectralGPT \cite{10490262}, FG-MAE \cite{wang2023feature},  msGFM \cite{han2024bridging}, \\ \, A2-MAE \cite{zhang20242}},  cause_leaf,  text width=34.0em
                        ]
                    ]
                    [Contrastive-Generative \\ Hybrid,  color=3!100,  fill=3_1!100,  text=black
                        [{\,  CMID \cite{10105625},  Cross-Scale MAE \cite{tang2024cross},  CROMA \cite{fuller2024croma},  MSFE \cite{10282433}},  cause_leaf,  text width=34.0em
                        ]
                    ]
                    [Geographic Knowledge,  color=3!100,  fill=3_1!100,  text=black
                        [{\,  GeoKR \cite{geokr_2022},  GeCo \cite{geco_2022},  MTP \cite{wang2024mtp},  GeRSP \cite{10400411},  TOV \cite{10110958},  SoftCon \cite{Wang2024MultiLabelGS},  SwiMDiff \cite{10453587},\\ \, CSPT \cite{rs14225675}, GFM \cite{Mendieta_2023_ICCV},  MMEarth \cite{nedungadi2024mmearth},  SARATR-X \cite{yang2024saratr}},  cause_leaf,  text width=34.0em
                        ]
                    ]
                ]
                [Multimodal Models,  color=4!100,  fill=4_1!100,  text=black
                    [Contrastive Learning,  color=4!100,  fill=4_1!100,  text=black
                        [{\,  RemoteCLIP\cite{remoteclip},  Skyscript\cite{skyscript},  GeoRSCLIP\cite{georsclip},  S-CLIP\cite{s-clip},  RS-CLIP\cite{rs-clip},  Mall et al.\cite{remoteWoAnn}, \\ \, Yang et al.\cite{bootstrapping},   GeoCLIP\cite{geoclip},  SatCLIP\cite{satclip},  AddressCLIP\cite{xu2024addressclip},  StreetCLIP\cite{haas2023learning},  GeoCLAP\cite{geoclap}},  cause_leaf,  text width=34.0em
                        ]
                    ]
                    [Auto Regressive Learning,  color=4!100,  fill=4_1!100,  text=black
                        [{\,  LHRS-Bot\cite{lhrs-bot},  H2RSVLM\cite{h2rsvlm},  Geochat\cite{geochat}, Skysensegpt\cite{skysensegpt},  RSGPT\cite{rsgpt},  SkyEyeGPT\cite{skyeyegpt},  \\ \,EarthGPT\cite{earthgpt},   Popeye\cite{popeye},  RS-CapRet\cite{rs-cap},  RSGPT\cite{rsgpt},  RS-LLaVA\cite{rs-llava}},  cause_leaf,  text width=34.0em
                        ]
                    ]
                ]
            ]
            [Training Dataset,  color=7!100,  fill=7!100,  text=black
                [Vision Models,  color=8!100,  fill=8!100,  text=black
                    [Custom Collected,  color=8!100,  fill=8!100,  text=black
                        [{\,  Seco \cite{Manas_2021_ICCV},  U-BARN \cite{10414422},  MATTER \cite{Akiva_2022_CVPR},  Levir-KR \cite{geokr_2022},  SkySense \cite{guo2024skysense},  GeoSense \cite{10378718},\\ \,  MMEarth \cite{nedungadi2024mmearth},   RingMo \cite{ringmo_2023},  TOV-RS \cite{10110958}}
                        , cause_leaf,  text width=34.0em
                        ]
                    ]
                    [Publicly Available,  color=8!100,  fill=8!100,  text=black
                        [{\, MillionAID \cite{millionaid},  SAMRS \cite{wang2024samrs},  Sen12MS \cite{sen12ms},  BigEarthNet-MM \cite{bigearthnetmm},  SSL4EO \cite{ssl4eo},\\ \,  SatlasPretrain \cite{satlaspretrain},  fMoW \cite{fmow},  fMoW-S2 \cite{satmae2022},  BigEarthNet-S2 \cite{sumbul2019bigearthnet},  MSAR \cite{chen2022large}, \\ \, SAR-Ship \cite{wang2019sar},  SARSim \cite{ kusk2016synthetic},  SAMPLE \cite{lewis2019sar}}
                        , cause_leaf,  text width=34.0em
                        ]
                    ]
                ]
                [Multimodal Models,  color=9!100,  fill=9!100,  text=black
                    [VLM,  color=9!100,  fill=9!100,  text=black
                        [{\, UCM-Caption \cite{qu2016deep},  RSICD \cite{lu2017exploring},  RSITMD \cite{yuan2022exploring},  DOTA \cite{dota},  DIOR \cite{dior},  fMoW \cite{fmow},  \\ \, MillionAID \cite{millionaid},  UCM \cite{yang2010bag},  NWPU-RESISC45 \cite{cheng2017remote},  SkyScript \cite{skyscript},  RS5M \cite{georsclip}}
                        , cause_leaf,  text width=34.0em
                        ]
                    ]
                    [MLLM,  color=9!100,  fill=9!100,  text=black
                        [{\, UCM \cite{yang2010bag},  NWPU-RESISC45 \cite{cheng2017remote},  UCM-Caption \cite{qu2016deep},  Sydney-Caption \cite{qu2016deep}, RSICD \cite{lu2017exploring}, \\ \,   NWPU-Caption \cite{cheng2022nwpu},   RSITMD \cite{yuan2022exploring},  RSVQA-LR \cite{lobry2020rsvqa},  RSVQA-HR \cite{lobry2020rsvqa},  RSIVQA \cite{zheng2021mutual}, \\ \,   FloodNet \cite{rahnemoonfar2021floodnet},  DOTA \cite{dota},  DIOR \cite{dior},  FAIR1M \cite{fair1m},  fMoW \cite{fmow},  MillionAID \cite{millionaid},  RSVG \cite{sun2022visual}, \\ \,   DIOR-RSVG \cite{zhan2023rsvg},  RSICap \cite{rsgpt},  LHRS-Align \cite{lhrs-bot},  LHRS-Instruct \cite{lhrs-bot}}
                        , cause_leaf,  text width=34.0em
                        ]
                    ]
                ]    
            ]
            [Evaluation,  color=10!100,  fill=10!100,  text=black
                [Vision Models,  color=11!100,  fill=11!100,  text=black
                    [Downstream Tasks,  color=11!100,  fill=11!100,  text=black
                        [{\,  Scene Classification  (Section \ref{sec: task_vis_cls}),  Object Detection  (Section \ref{sec: task_vis_object}),  \\ \, Semantic Segmentation  (Section \ref{sec: task_vis_segmen}),  Change Detection (Section \ref{sec: task_vis_change})}
                        , cause_leaf,  text width=34.0em
                        ]
                    ]
                    [Evaluation Datasets,  color=11!100,  fill=11!100,  text=black
                        [{\,  NWPU-RESISC45 \cite{cheng2017remote},  EuroSAT \cite{helber2019eurosat},  AID \cite{xia2017aid},  Xview \cite{xview},  DIOR \cite{dior},  DIOR-R \cite{diorr},  FAIR1M \cite{fair1m}, \\ \,  DOTA \cite{dota},  SpaceNetv1 \cite{spacenet},  LoveDA \cite{loveda},  iSAID \cite{isaid},  Dyna.-Pla. \cite{dynamicearthnet},  Dyna.-S2 \cite{dynamicearthnet},\\ \,   OSCD \cite{daudt2018urban},  LEVIR \cite{chen2020spatial}}
                        , cause_leaf,  text width=34.0em
                        ]
                    ]
                ]
                [Multimodal Models,  color=12!100,  fill=12!100,  text=black
                    [Downstream Tasks,  color=12!100,  fill=12!100,  text=black
                        [{\, Scene Classification (Section \ref{sec: task_mllm_scene}), Image Captioning (Section \ref{sec: task_mllm_caption}),  \\ \, Visual Question Answering (Section \ref{sec: task_mllm_vqa}), Visual Grounding (Section \ref{sec: task_mllm_grounding}), \\ \, Cross-Modal Retrieval (Section \ref{sec: task_mllm_retrieval})}
                        , cause_leaf,  text width=34.0em
                        ]
                    ]
                    [Evaluation Datasets,  color=12!100,  fill=12!100,  text=black
                        [{\,  EuroSAT \cite{helber2019eurosat},  NWPU-RESISC45 \cite{cheng2017remote},  WHU-RS19 \cite{dai2010satellite},  AID \cite{isaid},  SIRI-WHU \cite{zhao2015dirichlet}, \\ \, UCM-Caption \cite{qu2016deep},   Sydney-Caption \cite{qu2016deep},  RSICD \cite{lu2017exploring},  NWPU-Captions \cite{cheng2022nwpu},  RSITMD \cite{yuan2022exploring}, \\ \, RSVQA \cite{lobry2020rsvqa}, FloodNet \cite{rahnemoonfar2021floodnet},   RSIVQA \cite{zheng2021mutual},  RSVG,  DIOR-RSVG \cite{zhan2023rsvg}}
                        , cause_leaf,  text width=34.0em
                        ]
                    ]
                ]  
            ]
        ]
    \end{forest}
    }
    \caption{Organization and Taxonomy of this article.}
    \label{figure:categorization_of_survey}
\end{figure*}
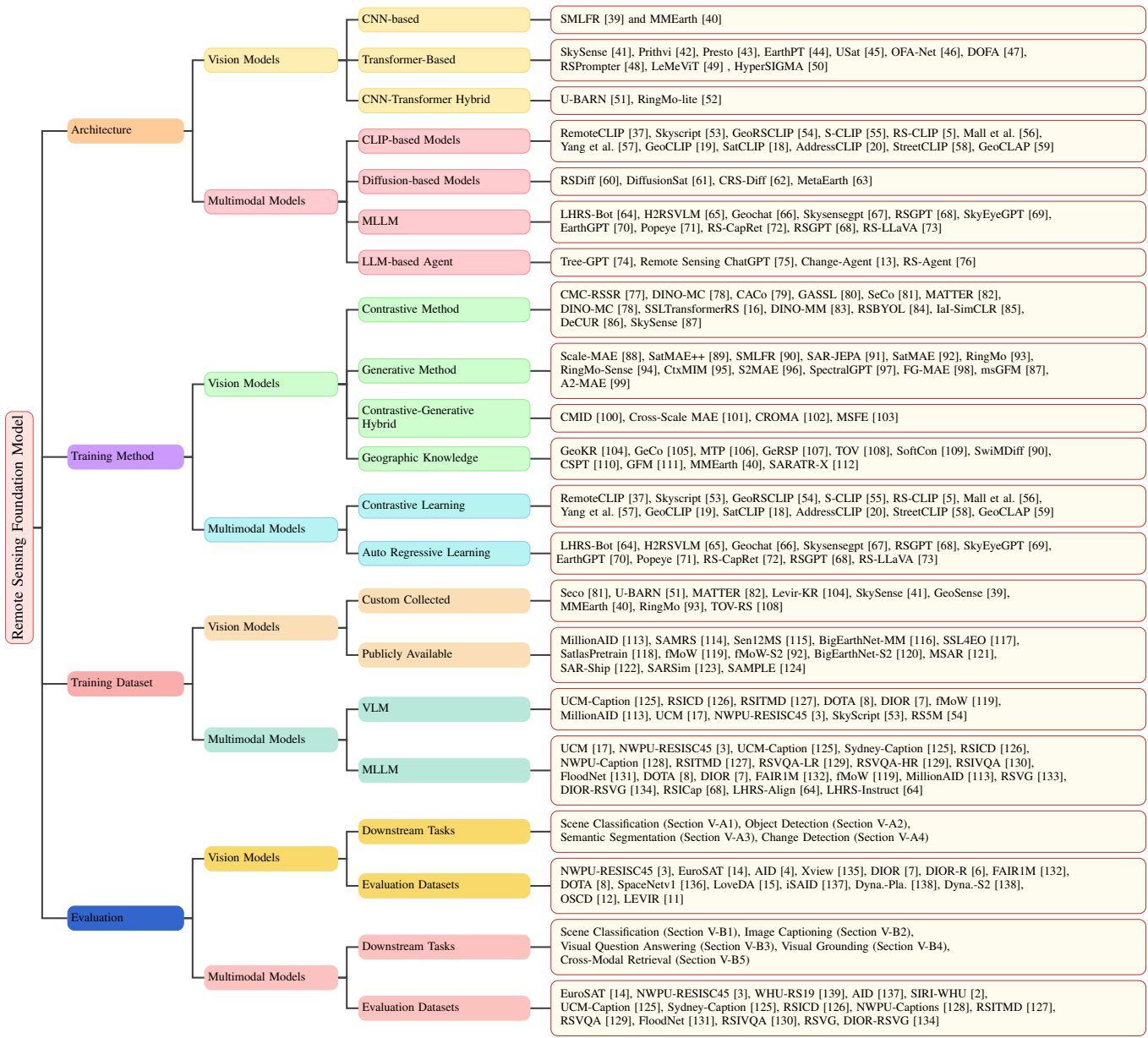

Model architecture, training methodologies, and datasets form the three fundamental pillars of deep learning, collectively shaping the generalization capabilities of trained models.
Recognizing their critical importance, this survey adopts a taxonomy centered on these three dimensions to systematically analyze the latest advancements in remote sensing foundation models.
By doing so, it offers a comprehensive reference for researchers striving to develop more effective and resilient remote sensing foundation models. 
Within this framework, we thoroughly examine the major improvement directions for vision and multimodal models, detailing the specific optimization strategies employed by each. 
Additionally, we synthesize the evaluation methodologies and results associated with existing foundational models to provide a holistic understanding of their strengths and limitations. 
The organization and taxonomy of this article is shown in Fig. \ref{figure:categorization_of_survey}. 
The \textbf{organization} of this review is as follows: Section \ref{sec:model} discusses model architectures; Section \ref{sec:training_methods} explores advancements in training methodologies; Section \ref{sec:training_data} delves into the construction and utilization of training datasets; Section \ref{sec:evaluation} presents evaluation benchmarks and corresponding results. Finally, we conclude by identifying key challenges and future research directions in Section \ref{sec:Challenges}. 

\section{Model Architecture}
\label{sec:model}

\subsection{Vision Foundation Model}

Deep learning techniques have showcased remarkable advancements in a multitude of remote sensing image processing tasks, encompassing image classification \cite{lu2007survey, rawat2017deep}, object detection \cite{szegedy2013deep, zhao2019object}, and semantic segmentation \cite{guo2018review, hao2020brief}, largely due to their robust ability to learn discriminative features. Nonetheless, these breakthroughs are predicated on the availability of extensive annotated datasets. The inherent multi-source and multi-resolution nature of remote sensing imagery poses substantial challenges. Consequently, when large-scale annotated remote sensing datasets are not accessible, models pretrained on natural images may yield sub-optimal performance when fine-tuned for subsequent tasks.

In response to this challenge, the concept of foundation models tailored for remote sensing images has emerged, aiming to harness the potential of vast amounts of unlabeled remote sensing data. These specialized models, as opposed to those pretrained exclusively on natural images, demonstrate enhanced proficiency in extracting pertinent features from remote sensing imagery, leading to more auspicious outcomes in downstream applications.

The architectural frameworks of these foundation models can be classified into three distinct categories: \textbf{CNN-based models} \cite{9284640, 10378718}, \textbf{Transformer-based models} \cite{guo2024skysense, tseng2024lightweight}, and \textbf{hybrid models} \cite{10414422, wang2023ringmolite} that integrate both CNN and Transformer components. An exhaustive examination of each of these categories will be provided in the subsequent sections.

\subsubsection{CNN-based Model} CNN was initially proposed in the 1980s \cite{GU2018354}, was inspired by the structure of the cat’s visual cortex \cite{hubel1962receptive}. 
In the pivotal year of 2012, the ImageNet competition \cite{krizhevsky2012imagenet} etched the CNN into the history of computer vision by achieving unprecedented accuracy in image classification. 
This breakthrough led to a surge in the adoption and development of CNNs, with subsequent innovations like ResNet \cite{he2016deep} in 2016 significantly enhancing model depth through residual connections, thereby pushing the boundaries of performance on large-scale natural image datasets.

Building on this momentum, early research in remote sensing \cite{Manas_2021_ICCV, kattenborn2021review} naturally leaned towards CNNs, with frameworks such as ResNet \cite{he2016deep} and YOLO \cite{jiang2022review} taking prominent roles. 
However, despite their considerable potential, inherent architectural constraints and limited parameter capacity presented significant challenges in scaling CNN-based approaches to process large-scale datasets.

To improve the performance of CNNs on extensive data, researchers have introduced architectures such as ConvNeXt \cite{9879745} and ConvNeXtv2 \cite{Woo_2023_CVPR}. 
ConvNeXt is designed to integrate the design principles of the Vision Transformer (ViT) \cite{alexey2020image} into the ResNet architecture, resulting in improved scalability of CNN-based models on large datasets. 
Motivated by the success of Transformers in mask-based representation learning, ConvNeXtv2 has undergone a redesign utilizing sparse convolutions to better align with the Masked Autoencoder (MAE) framework. 
As a result, several endeavors have applied these architectures to the remote sensing domain, including SMLFR \cite{10378718} and MMEarth \cite{nedungadi2024mmearth}. 
SMLFR \cite{10378718} utilized ConvNeXt as its visual encoder and features a lightweight decoder composed of three successive decoder blocks and two upsampling layers. 
MMEarth \cite{nedungadi2024mmearth} employed the ConvNeXt V2 architecture, which leverages sparse convolutions \cite{choy20194d} to improve efficiency. 
Both approaches have demonstrated competitive results when compared to Transformer-based methodologies. 
These findings emphasize the promise of foundation models employing ConvNeXt as their backbone for advancing remote sensing foundation models.

\subsubsection{Transformer-based Model} The Transformer architecture has surged in popularity in recent years, driven by its impressive scalability across both model size and dataset volume. 
ViT \cite{alexey2020image} extend the original transformer design \cite{vaswani2017attention} to image processing by treating image patches as token sequences. 
Traditional ViTs maintain a fixed number of tokens and token feature dimensions throughout the network, which limit performance. 
To overcome this, dense prediction models like PVT \cite{wang2021pyramid} and Swin Transformer \cite{liu2021swin} introduce multi-scale architectures, excelling in fine-grained tasks such as small object detection and segmentation. Moreover, the success of transformer-based models across various domains \cite{devlin2018bert, he2022masked, kirillov2023segment} highlights their advantage over CNNs for large-scale pretraining, making Transformer-based architectures a natural choice for many remote sensing image foundation models.

In the realm of remote sensing, most foundation models \cite{wang2022selfsupervised, rs14225675} adopt ViT or Swin Transformer as their backbone. 
Several studies also explore advanced ViT variants to further enhance performance. 
For instance, SARATR-X \cite{yang2024saratr} employed HiViT \cite{zhang2023hivit}, which merges Swin Transformer’s strengths with support for patch dropping to facilitate masked image modeling. 
LeMeViT \cite{jiang2024lemevit} integrated learnable meta tokens to efficiently compress image representations with a minimal set of learnable tokens. Building on the success of SAM, 
RSPrompter \cite{chen2023rsprompter} adopted prompt learning strategy in SAM \cite{sam} to generate semantically distinctive segmentation prompts tailored to remote sensing imagery. 
EarthPT \cite{smith2024earthpt} adapted the GPT-2 framework, replacing traditional word embeddings with multilayer perceptrons to effectively encode non-text data, expanding its versatility for remote sensing tasks.

To advance Transformer-based models, researchers have increasingly explored multi-branch architectures to enhance feature diversity, effectively tackling challenges such as multimodal fusion and fine-grained detail extraction.
Given the inherent differences between modalities, utilizing a single encoder to process features from diverse sources remains a significant challenge.
To address this limitation, several multimodal foundation models \cite{fuller2024croma, wang2023decurdecouplingcommon, guo2024skysense, Scheibenreif_2022_CVPR} employ modality-specific encoders, integrating the extracted features at a later stage.

Beyond modality separation, many methods leverage multi-branch designs to extract more intricate and complementary features. 
CtxMIM \cite{zhang2024ctxmimcontextenhancedmaskedimage} introduced a context-enhanced branch and a reconstructive branch to mitigate the issue of context missing. 
This design separates spatial feature extraction from feature fusion, enabling the integration of clues from modality, time, and geographic context. 
Furthermore, RS-DFM \cite{wang2024rs} introduced a dual-branch information compression module that is designed to separate high-frequency and low-frequency features. This approach facilitates efficient feature-level compression while preserving essential task-agnostic information. 
BFM \cite{10531642} explored a parallel configuration for multi-headed self-attention and feed-forward networks to improve performance for vision-related tasks, especially those requiring fine-grained spatial understanding, such as object detection and segmentation.


Moreover, numerous methods have specifically tailored components of the Transformer, like attention mechanism and patch encoding, to more accurately capture pertinent geographical features.

The attention mechanism serves as the cornerstone of the Transformer architecture. 
Various approaches endeavor to refine the attention mechanism within the remote sensing foundation model in order to enhance feature extraction or optimize performance. 
In terms of feature extraction, HyperSIGMA \cite{wang2024hypersigma} introduced an innovative sparse sampling attention mechanism designed to tackle spectral and spatial redundancy challenges in hyperspectral images. 
This mechanism enables the extraction of diverse contextual features and serves as the core component of HyperSIGMA, aimed at addressing the limitations in feature utilization for hyperspectral images. 
Despite their rich spectral information, hyperspectral images have traditionally been constrained to narrow, task-specific applications. 
RingMo-Aerial \cite{diao2024ringmo} proposed frequency-enhanced multi-head self-attention to address the issues of multi-scale variations and occlusions that result from oblique angles in remote sensing imagery. 
Regarding efficiency, RVSA \cite{rvsa_2023} introduced a novel rotated varied-size window attention mechanism as a replacement for the traditional full attention in transformers, substantially decreasing computational overhead and memory consumption. 
Simultaneously, it augments object representation by deriving rich contextual information from the diverse windows it produces. 
LeMeViT \cite{jiang2024lemevit} presented dual cross attention to enable seamless information exchange between image tokens and meta-tokens within this framework, resulting in a substantial reduction in computational complexity when compared to self-attention mechanisms.

The Transformer-based model is required to convert the image into patch embedding and then input it into the subsequent transformer module. 
For multimodal models, multiple independent patch embedding layers are generally utilized to generate patch embeddings of different modalities \cite{xiong2024all, han2024bridging}. 
DOFA \cite{xiong2024neural} introduced a wavelength-conditioned dynamic patch embedding layer to unify the input of various earth observation modalities. 
Thus, a unified network architecture can be trained across diverse data modalities. 
SpectralEarth \cite{braham2024spectralearth} utilized 4$\times$4 patches instead of the standard 16$\times$16 patches, preserving fine spatial details and enhancing the retention of spectral information at the patch projection layer.

Positional encoding plays a crucial role in Transformer-based models, providing spatial and structural context to the input data. 
Many studies in remote sensing foundation models adjust the positional embedding to adapt to pretraining on remote sensing images. 
SatMAE \cite{satmae2022} introduced positional encoding for the temporal/spectral dimension and independently masks patches across the temporal/spectral dimension, which allows the model to learn representations of the data that are more conducive to finetuning. 
Scale-MAE \cite{10377166} extended the positional encoding to include ground sample distance (GSD) by scaling the positional encoding relative to the land area covered in an image. 
Prithvi \cite{jakubik2023foundationmodelsgeneralistgeospatial} introduced 3D positional embeddings and 3D patch embeddings into the ViT framework to enable the model to handle spatiotemporal data. 
USat \cite{irvin2023usat} modified the patch projection layer and positional encoding to model spectral bands of different spatial scales from multiple sensors. 
This approach substantially reduces sequence length, and consequently memory footprint and runtime, while preserving the geospatial alignment of images from different sensors. 
To give the model more dimensional perception capabilities, MA3E \cite{li2024ma3e} added angle embeddings to the patch, enabling the model to perceive the angle of the patch.

\subsubsection{CNN-Transfromer Hybrid Model} 
To leverage the complementary strengths of CNNs and Transformers, recent approaches integrate both architectures by utilizing CNNs for efficient local feature extraction and Transformers for capturing global context and long-range dependencies.
U-BARN \cite{10414422} integrated U-Net and Transformer architectures to process the spatial, spectral, and temporal dimensions of data, effectively capturing the spatiotemporal information embedded in irregularly sampled multivariate satellite image time series.
The intricate interplay between high-frequency and low-frequency spectral components in remote sensing imagery limits the effectiveness of conventional CNNs and ViTs. 
RingMo-Sense \cite{10254320} employed the Video Swin Transformer as its backbone and utilizes a transposed convolution layer to upsample features spatially and temporally. 
SatMAE++ \cite{noman2024rethinking} utilized a CNN-based contract-upsample block to upsample the spatial resolution of the features for the multi-scale reconstruction procedure. 
RingMo-lite \cite{wang2023ringmolite} leveraged Transformer modules as low-pass filters to extract global features of remote sensing images through a dual-branch structure, combined with CNN modules as stacked high-pass filters to effectively capture fine-grained details. 
This successful implementation leads to a lightweight network that achieves excellent performance across various downstream remote sensing tasks. 
OmniSat \cite{astruc2024omnisat} utilized a CNN-based encoder-decoder for image processing and a lightweight temporal attention encoder \cite{lightweighttemporalattention} for time-series data, aligning with the inherent characteristics of data.

\subsection{Multimodal Foundation Models}

\begin{figure*}[t]
\centering
\includegraphics[width=\linewidth,scale=0.7]{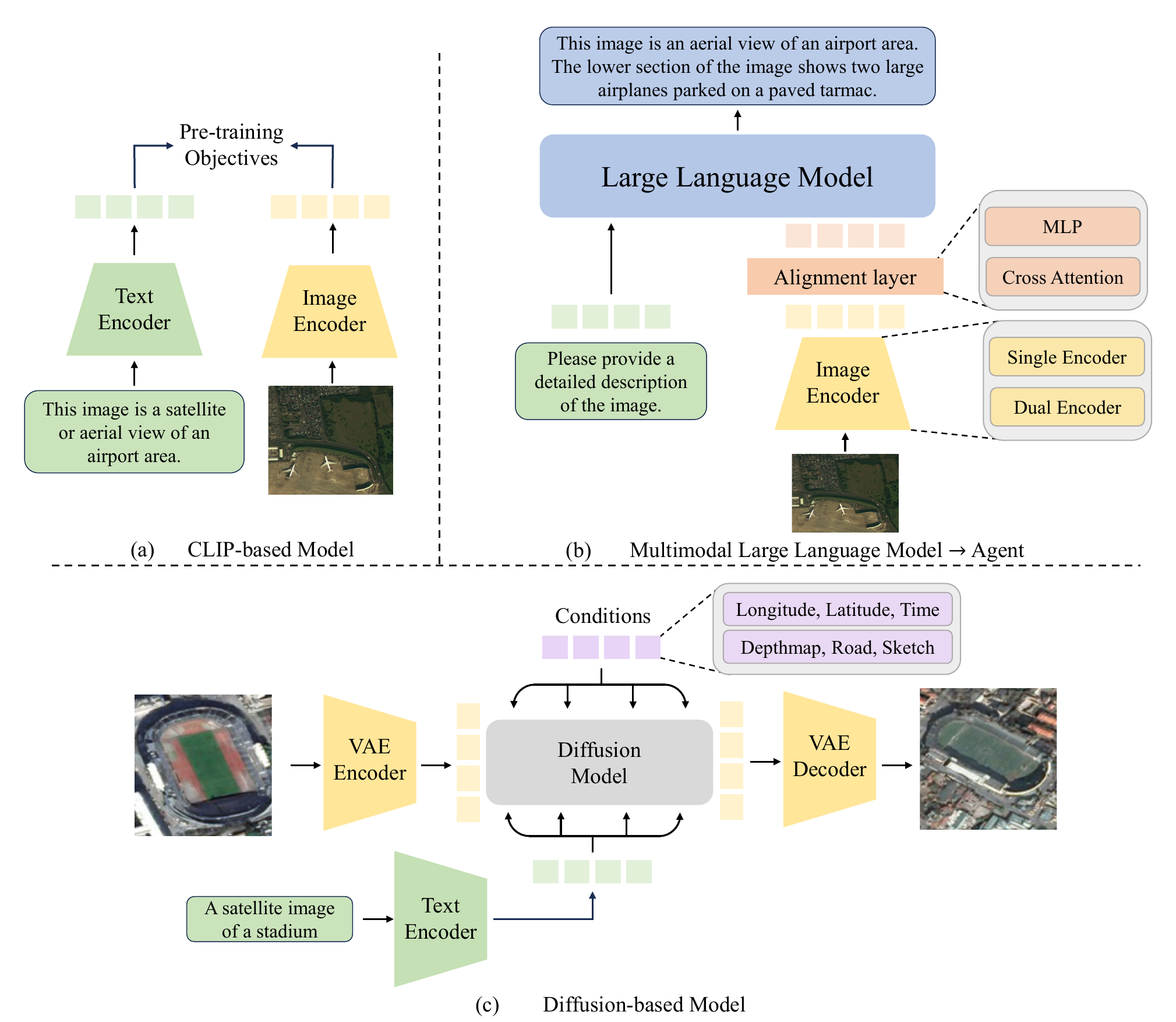}
    \caption{The architectures of remote sensing multimodal foundation models, primarily includes CLIP-based model, MLLM, and diffusion-based model. Agent models are built upon the LLM architecture.}
    \label{figure:multimodal-fig}
\end{figure*}

Recent research prioritize leveraging large-scale datasets and immense computational power to address a wide array of challenges. 
The primary objective is to develop a unified model that can handle various modalities and tasks simultaneously. 
As a result, there is an increasing focus on building and applying multimodal models, which are designed to integrate different types of data and perform well across multiple tasks. 

VLMs have been the focus of intensive research due to the capability of zero-shot predictions  across various visual recognition tasks, learning rich vision-language correlations from web-scale image-text pairs, which are abundantly available online. 
CLIP\cite{clip} is a seminal VLM with powerful zero-shot prediction capabilities, inspiring researchers to continually refine and innovate its subsequent iterations. 
These advancements encompass improvements in long-text capabilities\cite{zhang2024long}, multimodal feature fusion methods\cite{yu2022coca}, and model architecture\cite{wang2023image, alayrac2022flamingo}, among other areas.

With the rapid development of LLMs, multimodal large language models (MLLMs), exemplified by GPT-4V \cite{achiam2023gpt}, have emerged as a new and rapidly growing research focus. 
By harnessing the powerful reasoning capabilities of LLMs, MLLMs are able to perform tasks beyond the scope of traditional multimodal models. 
The subsequent research expanded the use cases and capabilities of MLLMs, including the extension of input and output types\cite{han2023imagebind, moon2023anymal}, such as video and point clouds, enhancement of the model's fine-grained processing abilities\cite{chen2023shikra, yuan2024osprey}, and increased language support\cite{hu2023large, bai2023qwen}.

The multimodal foundation model in the field of remote sensing can be roughly divided into the following four categories based on their structure: CLIP-based model (in Fig. \ref{figure:multimodal-fig}(a)), Multimodal Large Language Model (in Fig. \ref{figure:multimodal-fig}(b)), Diffusion-based model (in Fig.\ref{figure:multimodal-fig}(c)), and Agent.

\subsubsection{CLIP-based Model} Although vision foundational models have achieved fantastic results in various downstream tasks, fine-tuning is still necessary for specific downstream applications. 
Inspired by recent breakthroughs in NLP, VLM, which trained on large-scale image-text datasets and can be applied directly to downstream visual recognition tasks without requiring further fine-tuning, has gained significant attention. 
CLIP\cite{clip} employs pretraining with a large amount of image-text paired data for alignment, enabling zero-shot predictions in downstream tasks, such as image classification, cross-modal retrieval. 
In the field of remote sensing, most CLIP-based studies\cite{s-clip,remoteclip,georsclip,skyscript,rs-clip, xu2024addressclip} follow the same model structure as CLIP, as illustrated in Fig. \ref{figure:multimodal-fig} (a). 
CLIP\cite{clip} is a dual-tower model consisting of an image encoder and a text encoder. 
The image encoder, which can be either a ResNet or a ViT, converts images into visual embeddings. 
The text encoder, based on a Transformer architecture, processes sequences of word tokens and generates a vectorized representation. 

Mall et al.\cite{remoteWoAnn} leveraged CLIP's vision-language alignment capability and utilizes a dual-image encoder structure to spatially align remote sensing images with ground-level internet images taken at the same location, resulting in a vision-language foundation model for remote sensing images. 
Due to the significant differences in spatial resolution of remote sensing images, it is challenging to align the features of remote sensing images and text. 
Li et al. \cite{bootstrapping} introduced a lightweight interactive Fourier transformer module for remote sensing image captioning task. 
The module comprises a shared-parameter Fourier-based image Transformer and a Fourier-based text Transformer, which extract multi-scale features of remote sensing images in the frequency domain, enhancing the alignment of image and text features during the two-stage pretraining process.
The learnable visual prompts are input into the Fourier-based image Transformer, where they interact with the features extracted from the frozen image encoder to capture multi-scale visual information. Concurrently, the original text data is processed by the Fourier-based text Transformer for text feature extraction.

By aligning images with textual descriptions, CLIP has demonstrated remarkable generalization capabilities, inspiring studies that extend this paradigm to other modalities in remote sensing.
CSP \cite{csp} introduced a vision-location self-supervised learning framework employing a dual-encoder architecture that independently encodes images and location information, achieving alignment through contrastive learning.
GeoCLIP \cite{geoclip} introduced a foundational model for image geo-localization tasks, aligning pretrained CLIP image encoders with GPS coordinates. 
It encoded GPS coordinates using random Fourier features and employed an exponential sigma assignment strategy to facilitate hierarchical feature learning across different resolutions.
Another contemporaneous work, SatCLIP\cite{satclip}, is a task-agnostic, globally covering location encoder that matches globally-distributed satellite images with their corresponding coordinates using contrastive learning. 
They utilized the location encoder proposed by \cite{russwurm2023geographic}, which use spherical harmonics basis functions as positional encoders and combined with sinusoidal representation networks. 
GEOCLAP \cite{geoclap}, built upon CLIP, encodes three modalities: geo-tagged audio recordings, textual descriptions of audio, and overhead imagery of corresponding locations. 
It employed contrastive learning to align these modalities within a unified embedding space.

\subsubsection{Diffusion-based Model} 
Diffusion models are generative models that progressively destruct data by introducing noise into data, then learn to reverse this process. 
The structure of a typical diffusion model is depicted in Fig. \ref{figure:multimodal-fig} (c), consisting of two encoders for text and image and a conditional diffusion model. 
Conditional diffusion models extend the core principles of diffusion processes by integrating conditional information, enabling the generation of data that conforms to specific constraints.
In remote sensing, the limited availability of high-quality data highlights the necessity of developing advanced Diffusion-based models.

RSDiff \cite{sebaq2023rsdiff} proposed a novel and lightweight framework consisting of two cascading diffusion models to generate high-resolution satellite images from text prompt. 
DiffusionSat \cite{diffusionsat} utlized commonly associated metadata with satellite images including latitude, longitude, timestamp to train the model for single-image generation. 
A 3D control signal conditioning module, which is capable of processing a sequence of images, was designed to generalize to inverse problems such as multispectral super-resolution, temporal prediction, and in-painting. 
CRS-Diff \cite{Crs-diff} proposed the first multi-condition controllable generative foundation model for remote sensing.
It incorporated ControlNet \cite{zhang2023adding} to integrate two additional control signals into the diffusion model, enabling controlled RS image generation by refining both global and local conditions.
The model leverages six additional image control conditions (e.g., semantic segmentation masks, roadmaps, and sketches) alongside textual conditions (e.g., prompts, content images, and metadata encoding). 
MetaEarth \cite{metaearth} is a generative foundation model tailored for global-scale remote sensing image synthesis. 
By analyzing generation conditions and initial noise, it introduced a novel noise sampling strategy for denoising diffusion models, ensuring stylistic and semantic consistency across generated image tiles. 
This approach enables the seamless generation of arbitrarily large images from smaller components.

\subsubsection{Multimodal Large Language Model} 
Recently, with the continuous emergence of large language models, MLLMs have also developed rapidly and achieved great success in various vision-language tasks such as image description and visual question answering. 
MLLM refers to the LLM-based model capable of receiving, reasoning, and generating outputs based on multimodal information. Since the released GPT-4\cite{achiam2023gpt} demonstrates its impressive multimodal capabilities, the MLLM field has made rapid progress and shown promising results in different scenarios. 
However, due to significant differences in imaging conditions and scales between natural images and remote sensing images, general MLLMs do not perform well in the remote sensing field. 
Consequently, several MLLM studies specifically targeting the remote sensing domain have emerged \cite{rsgpt,skyeyegpt,rs-llava,skysensegpt,h2rsvlm}. 
These models adopt the architecture of existing MLLMs, comprising an image encoder, an alignment layer, and an LLM, as shown in Fig. \ref{figure:multimodal-fig}. 

The vision encoder is used to extract visual features from images, functioning as an external sensor similar to the human eye. 
All models employ ViT as their visual encoder, with pretrained weights sourced from either CLIP \cite{clip} or EVA-CLIP \cite{sun2023eva}. 
These pretrained models are usually already aligned with the text modality, and using such models to align with LLMs can be beneficial. In addition, there is a significant amount of exploration into different types of vision encoder. 
Osprey \cite{yuan2024osprey} incorporated a convolution-based ConvNext-L encoder \cite{cherti2023reproducible}, designed to leverage higher resolution inputs and extract multi-level features more effectively. 
Fuyu \cite{fuyu-8b} explored an encoder-free architecture, which project the image patches directly before being passed to the LLMs. 
Recent researches have also focused on enhancing image resolution to improve the model's perceptual capabilities. 
CogAgent \cite{hong2024cogagent} introduced a dual-encoder mechanism in which two separate encoders process high- and low-resolution images. 
The patch-division method segments a high-resolution image into patches and reuses the low-resolution encoder.
Monkey \cite{li2024monkey} and SPHINX \cite{lin2023sphinx} divided large images into smaller patches, sending sub-images along with a downsampled high-resolution image to the image encoder. 
The sub-images and low-resolution image capture local and global features, respectively.

Alignment layer is responsible for bridging the gap between vision modality and language modality.
Two methods are typically employed in the alignment layer: one, based on BLIP2 \cite{li2023blip}, incorporates a Q-Former and a linear layer, while the other relies solely on a linear layer \cite{liu2023llava}. The Q-Former compresses the length of visual tokens by selecting the most informative tokens as input.

The LLM in these models is selected from the latest state-of-the-art, open-source LLMs. 
LLaMA series \cite{touvron2023llama, touvron2023llama2} and Vicuna family \cite{vicuna} are representative open-sourced LLMs that have attracted much academic attention. 
However, these models are limited in their support for multiple languages, particularly Chinese. 
In contrast, Qwen \cite{bai2023qwen} is a bilingual LLM that offers strong support for both Chinese and English. 
These models, extensively pretrained on vast web corpora, are embedded with rich world knowledge and demonstrate strong generalization and reasoning abilities. 
This foundation enables them to perform well across diverse tasks without the need for training from scratch, significantly reducing both time and computational resources.

GeoChat \cite{geochat} is a multi-task remote sensing MLLM capable of performing image-level dialogues as well as dialogues targeting specific regions within an image. 
They created a distinct task identifier for each vision-related task to enable the model to switch between various types of visual interpretation tasks. 
RS-CapRet \cite{rs-cap} is a model proposed for image captioning and text-image retrieval tasks, designing a special retrieval token [RET] to retrieve images based on the similarity between the image and the [RET] token. 
EarthGPT \cite{earthgpt} proposed a unified MLLM integrating various multi-sensor remote sensing interpretation tasks and introduces a visual enhancement perception mechanism to refine visual perception information by combining visual features from CNN and ViT backbones. 
Popeye \cite{popeye} proposed the first MLLM for multi-source ship interpretation and multi-granularity ship detection tasks, utilizing various pretrained vision backbones to obtain multi-scale image features. 
The features from different backbones are then concatenated and refined using linear layer projections. 
To address the complexity of landscapes and variability in visual scales in remote sensing images, 
LHRS-Bot\cite{lhrs-bot} adopted a new bridging strategy, a multi-level vision-language alignment strategy. Learnable queries, which applied a descending query allocation strategy, was designed for different levels of image features, retained from various layers of the image encoder. 

\subsubsection{LLM-based Agent} 
AI agents \cite{xi2023rise} are artificial entities equipped with sensors to perceive their environment, make decisions based on that perception, and take actions using actuators in response to the conditions around them. 
The impressive reasoning capabilities of large language models like GPT-4 have led to increasing attention on LLM-based agents in recent research. In these systems, the LLM functions as the brain equipped with various tools to perceive and take action. These LLM-based agents demonstrate reasoning and planning abilities by utilizing techniques such as Chain-of-Thought (CoT) and problem decomposition\cite{wei2022chain, kojima2022large, wang2022self, zhou2022least}. 
Additionally, they can develop interactive capabilities with the environment by learning from feedback and executing new actions\cite{akyurek2023rl4f, peng2023check, liu2023languages}. LLM-based agents have been deployed in various real-world applications, including software development\cite{li2023camel} and scientific research\cite{boiko2023emergent}. 
However, in the field of remote sensing, research on agents is still in the exploratory stage.

Tree-GPT \cite{treegpt} utilized GPT-4 as the execution agent and incorporates an image understanding module and a domain knowledge base. 
The image understanding module automatically or interactively generated prompts to extract structured information from forest remote sensing images, guiding SAM to produce forest segmentation results. 
The system then calculated tree structure parameters based on these results and stores them in a database. 
Remote Sensing ChatGPT \cite{remote-sense-chatgpt} integrated various remote sensing task models using ChatGPT to solve complex task interpretations. 
It generated specific prompt templates based on user instructions to help ChatGPT accurately understand the commands and utilized the BLIP model to add captions to the input images, thereby injecting visual information into ChatGPT. 
Change-Agent \cite{change-agent} proposed an agent designed for remote sensing image change detection, comprising the multi-level change interpretation (MCI) model and LLM. 
The MCI model adopted a dual-branch structure with a shared underlying layer to handle both pixel-level and semantic-level change detection tasks. 
The LLM, acting as the agent, was equipped with a set of Python tools, enabling it to autonomously write Python programs to invoke the vision extraction backbone, change detection branch, change caption branch, and other relevant Python libraries. 
RS-Agent \cite{rs-agent} equipped the MLLM with a solution searcher, which provides solution guidance to assist the LLM in selecting appropriate tools from the tool space, and a knowledge searcher, which offers knowledge guidance.

\section{Training Methods}
\label{sec:training_methods}

Remote sensing imagery presents unique challenges and opportunities for training VLMs. Unlike conventional images, remote sensing data often contains multi-spectral and high-dimensional information, large-scale spatial coverage, and temporal variability. These characteristics make it essential to develop training strategies that can effectively extract meaningful patterns while addressing the domain-specific issues of remote sensing, such as data heterogeneity, resolution variation, and the scarcity of annotated datasets.

Traditional self-supervised methods, such as contrastive learning and masked autoencoders (MAE), have shown promise in leveraging the unlabeled nature of remote sensing data. Contrastive learning aims to learn robust representations by distinguishing between similar and dissimilar pairs, which is particularly useful for remote sensing tasks where intra-class variance can be high due to environmental factors. Similarly, MAE leverages the inherent redundancy in remote sensing imagery, enabling models to reconstruct masked patches and learn spatial-semantic correlations. However, while these methods can train effective models, they often fall short of fully addressing the unique demands of remote sensing, such as the need for multimodal fusion or domain-specific understanding.

The evolution of training methodologies, moving beyond contrastive learning and MAE, has led to the emergence of multimodal pretraining frameworks. These methods combine visual and textual information, enabling models to bridge the gap between remote sensing imagery and natural language descriptions. Recent advances have further extended this paradigm into MLLM pretraining, where visual features from remote sensing data are aligned with language models to enhance multimodal understanding. This progression culminates in techniques like alignment with human preferences, where models are fine-tuned to generate outputs aligned with human judgments, ensuring greater applicability in real-world scenarios.

\subsection{Vision Foundation Model}
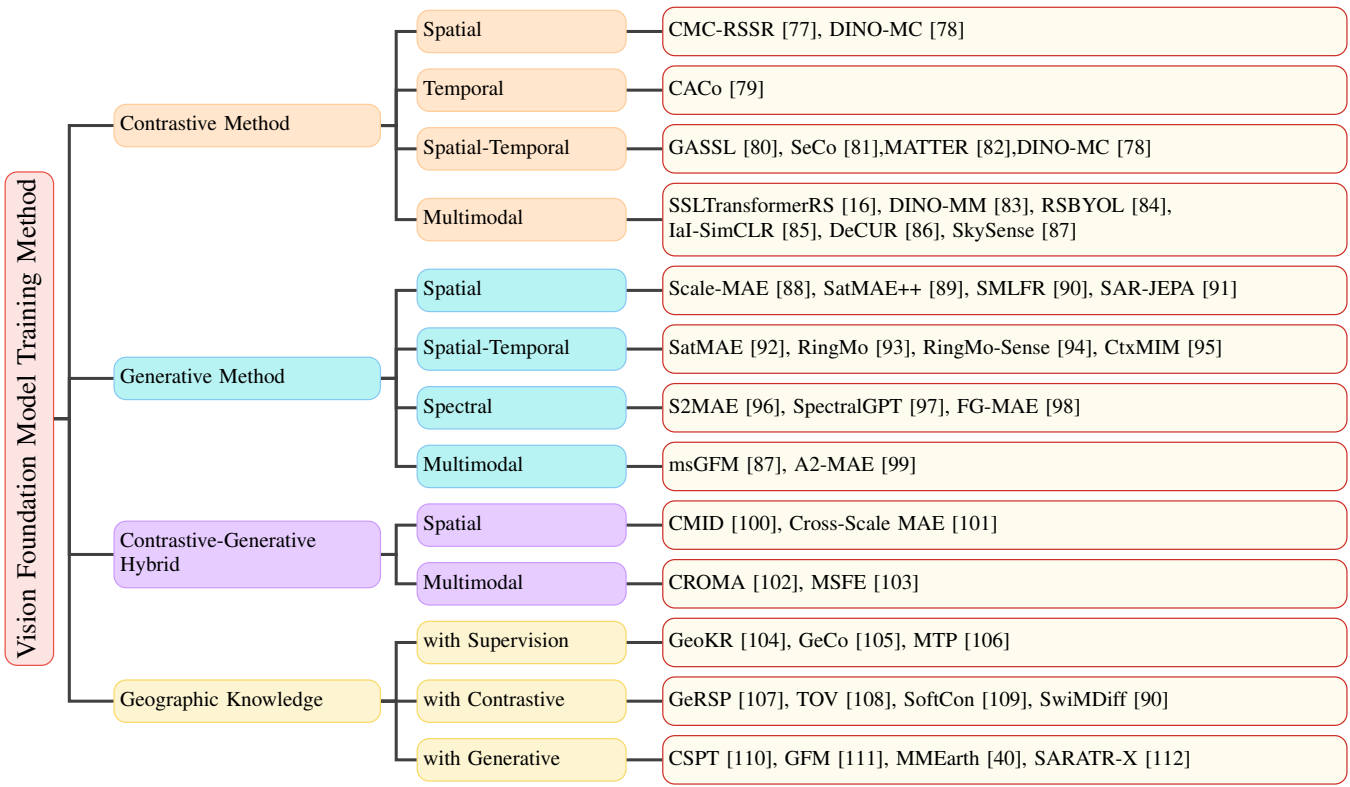
\begin{figure*}[t]
    \centering
    \resizebox{\textwidth}{!}{
    \begin{forest}
        forked edges,
        for tree={
            grow=east,
            reversed=true,
            anchor=base west,
            parent anchor=east,
            child anchor=west,
            base=left,
            font=\small,
            rectangle,
            draw=hidden-draw,
            rounded corners,
            align=left,
            minimum width=4em,
            edge+={darkgray, line width=1pt},
            s sep=3pt,
            inner xsep=2pt,
            inner ysep=3pt,
            ver/.style={rotate=90, child anchor=north, parent anchor=south, anchor=center},
        },
        where level=1{text width=8.4em,font=\scriptsize,}{},
        where level=2{text width=6.5em,font=\scriptsize,}{},
        where level=3{text width=22.2em,font=\scriptsize,}{},
        where level=4{text width=14.4em,font=\scriptsize,}{},
        [
           Vision Foundation Model Training Method, ver, color=carminepink!100, fill=carminepink!15, text=black
            [
                Contrastive Method, color=5!100, fill=5!50, text=black
                [
                    Spatial, color=5!100, fill=5!50, text=black
                    [
                        {CMC-RSSR \cite{Stojnic_2021_CVPR}, DINO-MC \cite{Wanyan_2024_CVPR}}, cause_leaf
                    ]
                ]
                [
                    Temporal, color=5!100, fill=5!50, text=black
                    [
                        {CACo\cite{Mall_2023_CVPR}}, cause_leaf
                    ]
                ]
                [
                    Spatial-Temporal, color=5!100, fill=5!50, text=black
                    [
                        {GASSL \cite{Ayush_2021_ICCV}, SeCo \cite{Manas_2021_ICCV},MATTER \cite{Akiva_2022_CVPR},DINO-MC \cite{Wanyan_2024_CVPR}}, cause_leaf
                    ]
                ]
                [
                    Multimodal, color=5!100, fill=5!50, text=black
                    [
                        {SSLTransformerRS \cite{Scheibenreif_2022_CVPR}, DINO-MM \cite{wang2022selfsupervised}, RSBYOL \cite{rs_byol_2022}, \\ IaI-SimCLR \cite{Prexl_2023_CVPR}, DeCUR \cite{wang2023decurdecouplingcommon}, SkySense \cite{han2024bridging}}, cause_leaf
                    ]
                ]
            ]
            [
                Generative Method, color=4!100, fill=4_1!100, text=black
                [
                    Spatial, color=4!100, fill=4_1!100, text=black
                    [
                        {Scale-MAE \cite{10377166}, SatMAE++ \cite{noman2024rethinking}, SMLFR \cite{10453587}, SAR-JEPA \cite{li2024predictinggradientbetterexploring}}, cause_leaf
                    ]
                ]
                [
                    Spatial-Temporal, color=4!100, fill=4_1!100, text=black
                    [
                        {SatMAE \cite{satmae2022}, RingMo \cite{ringmo_2023}, RingMo-Sense \cite{10254320}, CtxMIM \cite{zhang2024ctxmimcontextenhancedmaskedimage}}, cause_leaf
                    ]
                ]
                [
                    Spectral, color=4!100, fill=4_1!100, text=black
                    [
                       {S2MAE \cite{Li_2024_CVPR}, SpectralGPT \cite{10490262}, FG-MAE \cite{wang2023feature}}, cause_leaf
                    ]
                ]
                [
                    Multimodal, color=4!100, fill=4_1!100, text=black
                    [
                        {msGFM \cite{han2024bridging}, A2-MAE \cite{zhang20242}}, cause_leaf
                    ]
                ]
            ]
            [
                Contrastive-Generative \\ Hybrid, color=6!100, fill=6!50, text=black
                [
                    Spatial, color=6!100, fill=6!50, text=black
                    [
                        {CMID \cite{10105625}, Cross-Scale MAE \cite{tang2024cross}}, cause_leaf
                    ]
                ]
                [
                    Multimodal, color=6!100, fill=6!50, text=black
                    [
                        {CROMA \cite{fuller2024croma}, MSFE \cite{10282433}}, cause_leaf
                    ]
                ]
            ]
            [
                Geographic Knowledge, color=harvestgold!100, fill=harvestgold!30, text=black
                [
                    with Supervision, color=harvestgold!100, fill=harvestgold!30, text=black
                    [
                        {GeoKR \cite{geokr_2022}, GeCo \cite{geco_2022}, MTP \cite{wang2024mtp}}, cause_leaf
                    ]
                ]
                [
                    with Contrastive, color=harvestgold!100, fill=harvestgold!30, text=black
                    [
                        {GeRSP \cite{10400411}, TOV \cite{10110958}, SoftCon \cite{Wang2024MultiLabelGS}, SwiMDiff \cite{10453587}}, cause_leaf
                    ]
                ]
                [
                    with Generative, color=harvestgold!100, fill=harvestgold!30, text=black
                    [
                        {CSPT \cite{rs14225675}, GFM \cite{Mendieta_2023_ICCV}, MMEarth \cite{nedungadi2024mmearth}, SARATR-X \cite{yang2024saratr}}, cause_leaf
                    ]
                ]
            ]
        ]
    \end{forest}
    }
    \caption{Training methods for Remote Sensing Vision Foundation Models.}
    \label{figure:VisionTraining}
\end{figure*}

The robust generalization capability of vision foundation models originates from self-supervised learning on large-scale datasets. 
Most self-supervised approaches in remote sensing fall into contrastive and generative paradigms. 
Additionally, some methods seek to integrate both paradigms, leveraging their respective advantages. 
The organization of training methods for vision models is shown in Fig. \ref{figure:VisionTraining}. 

Contrastive methods learn more discriminative representations by comparing similar and dissimilar samples. Classical contrastive methods, such as MoCo \cite{he2020momentum}, SimCLR \cite{chen2020simple}, and BYOL \cite{grill2020bootstrap}, have demonstrated significant potential for representation learning. An increasing number of studies \cite{clip, sun2023eva, dinov2} have explored the integration of contrastive methodologies within large-scale pretraining frameworks, achieving remarkable successes.

With the success of ViT \cite{alexey2020image} and MAE \cite{he2022masked}, generative methods, particularly masked image modeling (MIM), have gained popularity for training vision foundation models. MIM learns image features through the approach of mask reconstruction. MIM methods, such as BEiT \cite{bao2021beit, peng2022beit} and SimMIM \cite{xie2022simmim}, have yielded promising results on large-scale data and are extensively utilized in diverse fields. 
Compared with contrastive methods, which often emphasize the global view and overlook the internal structure of the image, MIM primarily focuses on local relations. Naturally, recent studies aim to integrate the two training frameworks to enhance the model's performance on both image-level tasks (e.g. classification) and pixel-level tasks (e.g. segmentation).


\begin{figure*}[t]
\centering
\includegraphics[width=0.85\linewidth,scale=0.4]{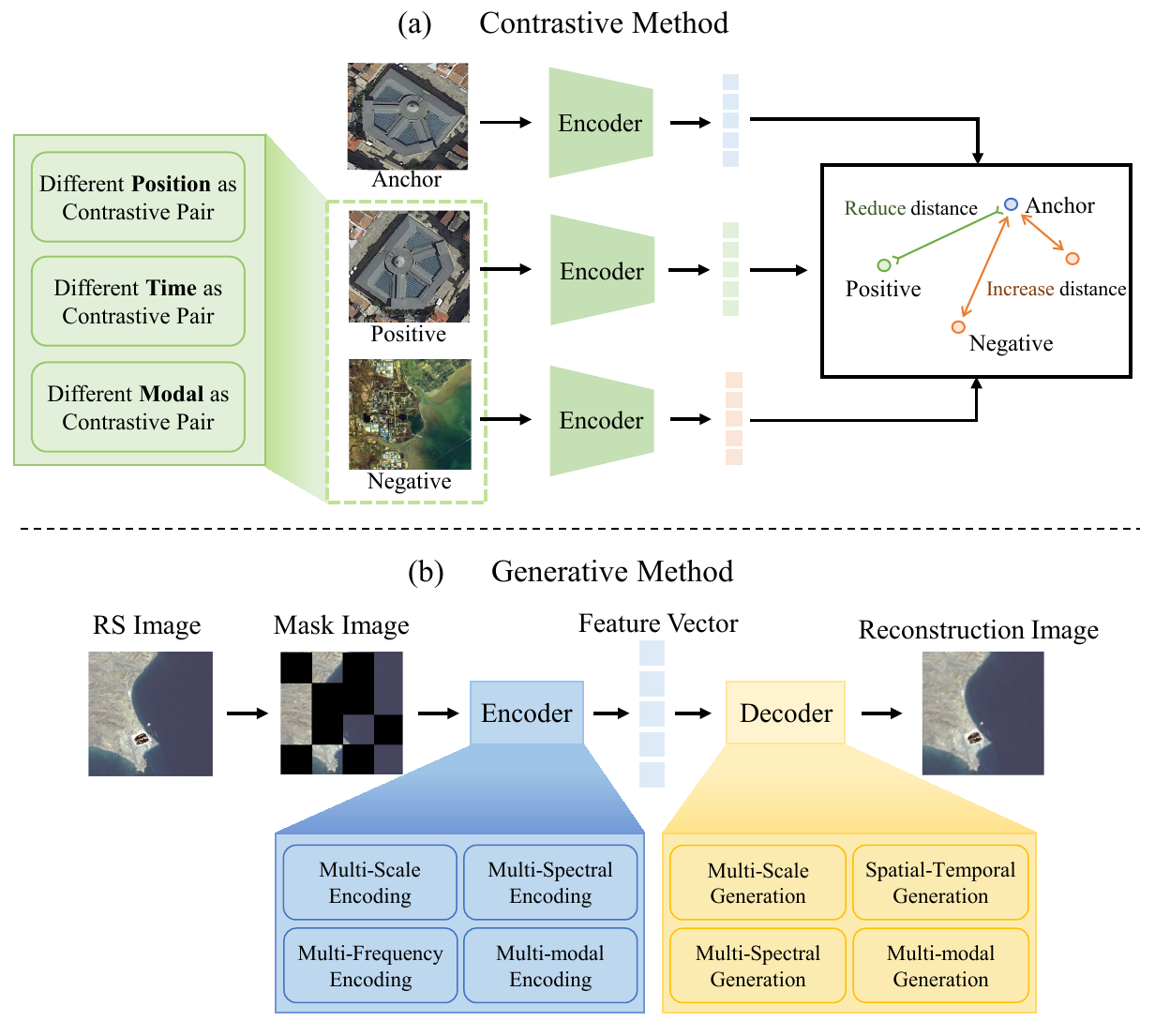}
    \caption{Contrastive and generative training methods for vision foundation model. Existing studies refine these methods by incorporating adaptations specific to remote sensing scenarios.}
    \label{figure:Fig_Visual_Models}
\end{figure*}

Subsequently, we will introduce how contrastive methods, generative methods, and contrastive-generative hybrid methods are applied in the field of remote sensing.

\subsubsection{Contrastive Method} Due to the modal differences between natural images and remote sensing images, it is challenging for models trained on natural images to attain satisfactory results on remote sensing tasks. Annotated remote sensing images are difficult to acquire due to the professional nature of remote sensing tasks. In addition to utilizing annotated training approaches, influenced by self-supervised contrastive methodologies like MOCO \cite{he2020momentum}, some methods have begun to explore the potential of pretraining on remote sensing images through contrastive learning.

The crucial element of contrastive learning is the methodology for constructing positive and negative sample pairs. 
For remote sensing images, the temporal and spatial contrast is of utmost importance. Numerous studies have constructed positive and negative sample pairs based on various factors, including location and season, as shown in Fig. \ref{figure:VisionTraining}. 
Furthermore, contrastive learning facilitates the construction of positive and negative samples across multiple modalities, capturing the shared attributes among them and enabling the model to process data from diverse sensors. We categorize the formulation of contrastive pairs for remote sensing images into four distinct categories: spatial contrast, temporal contrast, spatio-temporal contrast, and multimodal contrast, and will sequentially delve into each.


\lightblue{Spatial Contrast} For the spatial aspect, methods usually selects different views of the same image as positive samples to capture the intrinsic characteristics of remote sensing images. 
CMC-RSSR \cite{Stojnic_2021_CVPR} employed contrastive multiview coding which regard different views of the same picture as positive pairs and views of different images as negative pairs. 
This method enables the foundation model to achieve better results on downstream classification tasks compared to using models pretrained on natural scene images. 
Furthermore, DINO-MC \cite{Wanyan_2024_CVPR} employed the size variations of local crops to drive better representation learning of semantic content in remote sensing images.

\lightblue{Temporal Contrast} For the temporal aspect, methods introduce the concept of time dimension and use the difference in time to construct positive and negative samples. Mall et al. \cite{Mall_2023_CVPR} proposed a new change-aware contrastive loss, utilizing the contrast between temporal signals from long-term and short-term differences in images, as well as the relatively constant nature of satellite images. This approach allows the model to identify long-term permanent changes, such as house construction and lake desiccation.

\lightblue{Spatial-Temporal Contrast} 
Additionally, the concurrent utilization of temporal and spatial information enables the generation of more sophisticated contrast samples, thereby allowing the model to extract even richer and more nuanced features.
GASSL \cite{Ayush_2021_ICCV} utilized the spatio-temporal structure to construct positive and negative sample pairs through geo-aware contrastive learning, significantly narrowing the performance gap between MoCo-v2 and supervised learning on remote sensing images. 
SeCo \cite{Manas_2021_ICCV} developed a foundational model using multiple embedding subspaces for season contrast and space contrast.
The model exhibits superior performance on downstream tasks such as land-cover classification and change detection.
Akiva et al. \cite{Akiva_2022_CVPR} introduced MATTER, a contrastive method inspired by classical material and texture techniques. 
It utilized multi-temporal, spatially aligned remote sensing imagery over stable regions to ensure consistency in material and texture representation by achieving invariance to illumination and viewing angle.

\lightblue{Multimodal Contrast} 
Methods that focus solely on single-modality data do not fully utilize all available information. 
The multimodal contrastive learning methods establish positive sample pairs across different modalities, ensuring that the features possess the essential characteristics of each modality. 
Simultaneously, constructing negative samples within the same modality enhances the discriminative power of the features. 
Scheibenreif et al. \cite{Scheibenreif_2022_CVPR} utilized data from different sensors at the same location as positive samples and data from different locations as negative samples to train a foundational model through contrastive learning.
DINO-MM \cite{wang2022selfsupervised} built upon the DINO framework and enhanced input SAR-optical images through RandomSensorDrop, simultaneously extracting optical, SAR, and SAR-optical features.
To prevent multimodal contrastive learning from capturing only common features within modalities, RSBYOL \cite{rs_byol_2022} adopted BYOL as the foundational framework and trained RS-BYOL using single-channel and three-channel feature learning methods to embed invariant features across multispectral and synthetic aperture radar (SAR) sensors.

Prexl et al. \cite{Prexl_2023_CVPR} proposed Intra- and Inter-modality SimCLR (IaI-SimCLR), which encourages the model to simultaneously capture similarities between modalities and inherent features within each modality. 
DeCUR \cite{wang2023decurdecouplingcommon} constructed intermodality and intramodality comparison pairs, enabling it to capture both shared features across modalities and distinct features within individual modalities. 
SkySense \cite{guo2024skysense} proposed multi-granularity contrastive learning across various modalities and spatial granularities to process multimodal temporal remote sensing image sequences.

\subsubsection{Generative Method} 
Generative methods focus on obtaining representations by generating samples from the learned representation space \cite{khan2024surveyselfsupervisedlearning}.
Masked Image Modeling (MIM), where portions of an image are masked, and the model is trained to reconstruct them, is widely regarded as one of the most prevalent generative pretraining methods.
The majority of generative methods in remote sensing also adopt this approach. 
Different methods exhibit varying focuses. They are categorized into spatial generation, spatial-temporal generation, spectral generation, and multimodal generation, as shown in Fig. \ref{figure:VisionTraining}. 

\lightblue{Spatial Generation. 
}MIM selects patches at random for masking and then reconstructs the mask utilizing spatial context information. Given the high resolution, numerous small objects, and significant size variations within remote sensing images, the spatial generation method has been further refined to enhance multi-scale perception and small target detection, thereby enabling more effective feature extraction. 

Scale-MAE \cite{10377166} processed the masked image using a bandpass filter and reconstructed low-frequency/high-frequency images at lower/higher scales to obtain more robust multiscale representations.
SatMAE++ \cite{noman2024rethinking} employed multiscale pretraining strategies and incorporated upsampling blocks to reconstruct images at higher resolutions, facilitating the flexible integration of additional scales. 
This approach exhibits exceptional performance in both optical and multispectral satellite imagery.
To capture multiscale features and frequency-domain information, Dong et al. \cite{10378718} proposed the sparse modeling and low-frequency reconstruction (SMLFR) framework for self-supervised representation learning, allowing the model to process variable-length sequences while mitigating unnecessary detail interference. 
Training on SAR data presents challenges related to small-object loss and the pervasive noise inherent in SAR imagery. 
To address these challenges, SAR-JEPA \cite{li2024predictinggradientbetterexploring}, built upon the JEPA \cite{Assran_2023_CVPR} framework, enhancerf the extraction of contextual information around small targets through local masking and utilized multiscale gradient features as guiding signals to mitigate SAR speckle noise interference.

\lightblue{Spatial-Temporal Generation. }Remote sensing images inherently possess significant spatiotemporal characteristics, and generation methods must consider how to capture these characteristics during the generation process. 

Temporal SatMAE \cite{satmae2022} discussed the effects of various temporal masking methods on the generative model and discovered that independently masking each image while maintaining a consistent masking rate across images yielded optimal outcomes. 
Because the basic masking strategy may result in the loss of dense and small objects in remote sensing images, RingMo \cite{ringmo_2023} proposed a patch incomplete mask (PIMask) strategy that employs partially incomplete masks to preserve remote sensing features in complex scenes while maintaining the overall mask rate. 
To infuse characteristics with temporal continuity and spatial affinity, RingMo-Sense \cite{10254320} constructed a spatial, temporal, and spatio-temporal three-branch prediction network, utilizing block, tube, and frame masking methods, respectively, to acquire spatial affinity, temporal continuity, and spatiotemporal interaction. 
To mitigate the loss of context information, 
CtxMIM \cite{zhang2024ctxmimcontextenhancedmaskedimage} proposed a Siamese framework and fed masked and unmasked images into two separate branches.
The model extracts more comprehensive remote sensing features through the reconstruction of each branch while maintaining contextual consistency between them.

\lightblue{Spectral Generation. }Different from RGB images, spectral images often possess distinct features. 
Certain techniques have developed specialized algorithms tailored for spectral images. 
To specifically address spectral data, S2MAE \cite{Li_2024_CVPR} employed over 90\% 3D masking strategy. 
SpectralGPT \cite{10490262} adopted a multi-objective reconstruction strategy to comprehensively capture both local spatial-spectral features and spectral sequence information more effectively.
To enhance the reconstruction of multispectral and SAR features, FG-MAE \cite{wang2023feature} combined histograms of oriented gradients (HOG) and normalized difference indices (NDI) for multispectral images, while simultaneously reconstructing HOG for SAR imagery..

\lightblue{Multimodal Generation. }Remote sensing images are derived from a wide range of sources and display a variety of formats, including RGB, multispectral, and SAR, among others. 
Some studies simultaneously reconstruct multimodal remote sensing images, thereby enabling the model to support downstream tasks related to multiple modalities of remote sensing imagery.
OFA-Net \cite{xiong2024all} utilizes MAE to effectively fuse data representations from different modalities. 
MSGFM \cite{han2024bridging} implements a cross-sensor joint representation learning approach, wherein features from one sensor are employed to predict and reconstruct images from other sensors. This method fosters the alignment of representations across different sensors.
A2MAE \cite{zhang20242} introduces an anchor-aware masked autoencoder, which leverages intrinsic complementary information from various image types and geographic data to reconstruct masked patches during the pretraining phase.

\subsubsection{Contrastive-Generative Hybrid} Park et al. \cite{park2023what} discovered that combining contrast and reconstruction methods can result in obtaining better feature representation. In the realm of remote sensing, the contrastive-generative hybrid pretraining method has emerged as a promising approach in both single-modal and cross-modal scenes.

\lightblue{Spatial Enhancement. }
Muhtar et al. \cite{10105625} argued that representations learned solely through contrastive learning or masked generation methods have limited capacity to integrate global semantic separability and local spatial perception. 
To overcome this limitation, they introduced the Contrastive Mask Image Distillation (CMID) method, designed to learn both global semantics and local representations in a unified framework. 
To extract effective features from unaligned multi-scale images, Cross-Scale MAE \cite{tang2024cross} employed multi-scale contrastive learning and masked reconstruction to identify relationships across different scales.

\lightblue{Multimodal Enhancement. }
For multimodal remote sensing data, several studies have explored the use of generative methods to extract image features and leverage contrastive learning to integrate information across modalities. CROMA \cite{fuller2024croma} introduced a multi-vision modality framework that integrates contrastive and generative self-supervision objectives. It employed a generative paradigm to learn single-modal encoders and implemented cross-modal contrastive learning. Feng et al. \cite{10282433} incorporated patch-masking reconstruction loss and cross-modal data contrastive loss to address the issue of spatial coherence in heterogeneous modal features during cross-modal collaborative interpretation.

\subsubsection{Geographic Knowledge} Despite the significant challenges in annotating geographic data, numerous geographic data products and academic datasets still contain valuable annotations. To prevent the underutilization of these annotations, several methods have been employed to leverage them directly for model training or to enhance self-supervised training approaches, aiming to extract more accurate geographic features.

\lightblue{Geographic Knowledge with Supervision. }The simplest way to utilize geographic knowledge is to directly employ it to build a pretext task and train the foundation model. GeoKR \cite{geokr_2022} proposed a method that utilizes geographic knowledge as supervisory signals for learning remote sensing image representations. GeCo \cite{geco_2022} argued that previous methods using geographic supervision signals generated by GLC are biased. Therefore, they introduced a bias correction method for geographic supervision to improve the learning performance of remote sensing representations. Furthermore, as the amount of geographically annotated data increases, the latest methods attempt to use multiple datasets to train the model concurrently. Wang et al. \cite{wang2024mtp} performed multi-task pretraining on SAMRS and achieved promising results.

\lightblue{Geographic Knowledge with Contrastive. }To learn features that are aligned with the nature of remote sensing images, some methods have incorporated additional geographical knowledge into the training framework of contrastive methods. 

GeRSP \cite{10400411} incorporated natural images and classification supervision signals into the student model within the EMA framework, facilitating the learning of the characteristics of remote sensing images through visual knowledge. 
To enable the contrastive learning method to learn better geographical features, Tao et al. \cite{10110958} proposed an automatic data sampling and resampling mechanism leveraging geographic data products (such as OSM and FROM-GLC10) to build a large-scale, scalable, and relatively balanced remote sensing image dataset for training. 
SoftCon \cite{Wang2024MultiLabelGS} proposed soft contrastive learning that utilizes land cover-generated multi-label supervision to optimize cross-scene soft similarity, addressing multiple positive samples and strict positive matching in complex scenes. 
To mitigate the impact of similar samples in negative pairs found in previous contrastive learning methods, SwiMDiff \cite{10453587} improved contrastive learning by implementing a scene-wide matching strategy and enhanced the model's attentiveness to fine-grained features using a diffusion branch.

\lightblue{Geographic Knowledge with Generative. }
Some researched have focused on integrating remote sensing knowledge within the training methodologies of generative models. To explore the potential of multiple pretext tasks, 
Nedungadi et al. \cite{nedungadi2024mmearth} incorporated several auxiliary tasks within the MAE framework to facilitate training. Inspired by the "not stopping pretrain" concept in NLP,
Zhang et al. \cite{rs14225675} proposed consecutive pretraining, initially training on large-scale natural image data to identify general visual patterns, followed by pretraining on large-scale remote sensing images to familiarize the model with the semantics and content of task-specific image data. 
Mendieta et al. \cite{Mendieta_2023_ICCV} found that standard continuous pretraining offers only limited benefits. As a result, they introduced a method using MAE, which employs an ImageNet-pretrained model as a guiding model for feature reconstruction, efficiently leveraging the pretrained natural image model. 
Due to the speckle noise present in SAR images, which can interfere with pretext tasks, SARATR-X \cite{yang2024saratr} first pretrained the model by applying MIM on natural images and then on remote sensing images.

\subsection{Multimodal Foundation Model} 
Due to the gap between the text and visual modalities, the training method for multimodal models needs to align both modalities into the same space. 

VLM pretraining has been approached with three key objectives: contrastive, generative, and alignment. Contrastive learning plays a crucial role in VLM pretraining, where contrastive objectives are designed to capture distinguishing image-text features. 
Generative pretraining, by contrast, develops semantic understanding by generating images or text through techniques such as masked image modeling, masked language modeling, masked cross-modal modeling, and image-to-text generation. 
Alignment objectives ensure that VLMs correctly align image-text pairs by predicting whether a given text accurately describes its corresponding image.

MLLMs generally progress through the following training stages: pretraining, instruction tuning, and alignment tuning. Each phase of training requires different types of data and fulfills different objectives. The pretraining phase is aimed at aligning different modalities and providing world knowledge. Instruction tuning focuses on enhancing models' ability to comprehend user instructions and effectively execute the required tasks. In contrast, alignment tuning is typically applied in situations where it’s essential for models to align with particular human preferences.

Next, we will introduce the training methods of multimodal remote sensing foundation models by dividing them into three categories: contrastive learning, multimodal contrast, and auto-regressive learning.

\subsubsection{Contrastive Learning} The goal of image-text contrastive learning is to learn the correlation between visual and textual data by comparing image-text pairs, pulling the embeddings of paired images and texts closer together while pushing unrelated pairs farther apart. CLIP\cite{clip} employs a symmetrical image-text infoNCE loss, as shown in Eq. \ref{eq_img2text} and Eq. \ref{eq_text2img}, which measures the similarity between image and text embeddings using a dot-product. Most remote sensing VLM models\cite{remoteclip,skyscript,georsclip} are based on CLIP for continual pretraining. This involves initializing the CLIP model with weights pretrained on web-based image-text data, followed by further training on the remote sensing dataset using image-text contrastive learning. 

\begin{equation}
    \mathcal{L}_{I \rightarrow T}=- \frac{1}{B} \sum_{i=1}^B \log \frac{\exp{(z_i^I\cdot z^T_i/\tau)}}{\sum_{j=1}^{B}{\exp(z^I_i\cdot z^T_j/\tau)}},
    \label{eq_img2text}
\end{equation}
\begin{equation}
    \mathcal{L}_{T \rightarrow I}=- \frac{1}{B} \sum_{i=1}^B \log \frac{\exp{(z_i^T\cdot z^I_i/\tau)}}{\sum_{j=1}^{B}{\exp(z^T_i\cdot z^I_j/\tau)}},
    \label{eq_text2img}
\end{equation}

However, developing open-vocabulary vision-language models necessitates a vast amount of text-image pairs. This poses a significant challenge in the field of remote sensing. Unlike internet images, which are typically accompanied by captions or alt-text provided by their creators, satellite images are automatically captured by remote sensors with minimal human involvement and lack corresponding text annotations.

S-CLIP\cite{s-clip}, a semi-supervised learning method, introduced caption-level and keyword-level pseudo-label losses, which are then combined with CLIP's InfoNCE loss. 
The caption-level loss assumes that the semantics of an unlabeled image can be represented as a combination of the captions of labeled images. 
Keyword-level pseudo-label losses assumes that unlabeled images share keywords with visually similar images, even if their full captions are not identical. 
The pseudo-labels are defined as one of the keywords from the captions of the nearest labeled images, creating a candidate keyword set rather than a single exact label. 
RS-CLIP\cite{rs-clip} also utilized pseudo-labels for training and introduces a curriculum learning strategy, which presents training samples in a specific order or based on levels of complexity, allowing the model to learn more effectively. 
They utilized CLIP model to generate pseudo labels by selecting an identical number of samples for each class as pseudo labels, which prevents imbalanced distribution issues. 

Mall et al. \cite{remoteWoAnn} leveraged CLIP's alignment to spatially align remote sensing images with ground-level internet images taken at the same location, enabling the training of VLMs for remote sensing images without requiring any text annotations. 
Image-text contrastive learning was extended to both the image and pixel levels, addressing the challenge of a single satellite image corresponding to multiple ground images as well as the issue of pixel-level comprehension.

Yang et al.\cite{bootstrapping} proposed a two-stage pretraining framework. 
In the first stage, image-text contrastive learning was applied to align multi-scale image features with textual descriptions. 
A Fourier Transformer was used to extract visual features in the frequency domain, minimizing the gap between visual and textual modalities. In the second stage, the learned visual features guided a frozen language model using prefix causal language modeling, improving the quality of text generation.
This method effectively handles the multi-scale nature of remote sensing images, resulting in better image-text alignment and captioning.

\subsubsection{Multimodal Contrast}  Some studies have also extended this contrastive learning paradigm to remote sensing images and other modalities. 

GeoCLIP\cite{geoclip} and SatCLIP\cite{satclip} employed an image-location contrastive learning scheme to pretrained model. Simple address coordinates cannot provide rich semantic information, which increases the difficulty of aligning with image features. AddressCLIP\cite{xu2024addressclip} introduced image caption information as a supplement to the address text, promoting the alignment of images and addresses by using image-address contrastive loss and image-caption contrastive loss. 
Simultaneously, they proposed an image-geography matching mechanism, which utilizes spatial address distance to constrain the similarity of image features, ensuring that the distribution of image features closely aligns with geographical coordinates. 
StreetCLIP\cite{haas2023learning} employed synthetic caption pretraining to improve CLIP's zero-shot image geolocalization. By generating captions describing an image geographic location, it enables CLIP to associate visual content with geographic labels. 
GeoCLAP\cite{geoclap} was trained on embedding triplets using contrastive learning objective similar to CLIP\cite{clip} for all three pairs of embeddings, including the audio-text pair, the audio-image pair, and the image-text pair.

\subsubsection{Auto-regressive Learning} 
Auto-regressive learning involves predicting the subsequent word or character and is employed to train models for a broad spectrum of natural language tasks, including text generation, text classification, and question answering. This process generally consists of two stages: pretraining and instruction fine-tuning.

Pretraining primarily aims at aligning different modalities and acquiring multimodal world knowledge, typically utilizing large-scale text-paired datasets, such as caption data. These caption pairs generally describe images, audio, or videos in natural language sentences. Given an image, the model is trained to autoregressively predict the caption, following a standard cross-entropy loss.

Instruction tuning refers to the process of further training MLLMs on a dataset of instruction-output pairs in a supervised manner, bridging the gap between the next-word prediction objective of LLMs and the user's goal of having LLMs follow human instructions.

Given each sample as a list $X_{c} = (\boldsymbol{X}_{instruct}^{1}, \boldsymbol{X}_{a}^{1},... ,$ $ \boldsymbol{X}_{instruct}^{n}, \boldsymbol{X}_{a}^{n})$, where $\boldsymbol{X}_{instruct}^{n}$ is the instruction for $n$-th turn. For the given remote sensing image features $\boldsymbol{F}_{v}$, connect it with the instruction tokens $\boldsymbol{X}_{instruct}$ from the text modality. This concatenated input is then fed into the LLM. The model generates the answer $\boldsymbol{X}_{a}$ with a length of $L$. Maximizing the likelihood function that is defined as follows:

\begin{equation}
\label{sftloss}
\begin{split}
    \mathcal{L} & = \text{log}P \left (\boldsymbol{X}_{a} \mid \boldsymbol{F}_{v},\boldsymbol{X}_{instruct} ; \theta \right )   \\
    & = \sum_{i=1}^{L} \text{log}P \left ( x_{i} \mid \boldsymbol{F}_{v},\boldsymbol{X}_{instruct,<i},\boldsymbol{X}_{a,<i} ; \theta \right ),
\end{split}
\end{equation}

where $\boldsymbol{X}_{instruct,<i}$ is the instruction tokens in all turns before the current prediction tokens $x_{i}$. Therefore, the instructions and answers from previous rounds serve as references for the current task's response. Fine-tuning a MLLM or LLM is often challenging and computationally expensive due to the vast number of parameters. 

To mitigate this issue, parameter-efficient fine-tuning methods such as LoRA\cite{hu2021lora} have been introduced to finetune the pretrained models.  LoRA\cite{hu2021lora} is a widely adopted, parameter-efficient fine-tuning method due to its simplicity and effectiveness. Building upon the hypothes is that updates made during the fine-tuning exhibit a low intrinsic rank, the key concept of LoRA is the decomposition of the large weight matrix into two smaller matrices using low-rank decomposition. For a pretrained weight matrix $W^{0}\in \mathbb{R}^{d\times k}$, we define the parameter efficient fine-tuning as an update matrix $\Delta W$. LoRA \cite{hu2021lora} decomposes it into the product of two low-rank matrices:

\begin{equation}
\label{lora}
W=W^{0}+ \Delta W=W^{0}+BA ,
\end{equation}
where $ A \in\mathbb{R}^{r\times k}, B \in\mathbb{R}^{d\times r} $ and the rank $r\ll \min \left( d, k \right)$. The matrix A is initialized with uniform Kaiming distribution, while B is initially set to zero.

For large remote sensing multimodal models, we primarily divide the training methods into two categories for discussion: pretraining to fine-tuning and only fine-tuning. 

\lightblue{Pretraining to Fine-tuning. }
In some study, the model undergoes continued pretraining to align it with remote sensing images, followed by instruction fine-tuning to adapt the model to downstream tasks.

LHRS-Bot \cite{lhrs-bot} proposed a three-stage curriculum learning strategy to progressively align the features of the visual and language modalities.
In the first stage, the vision encoder was pretrained using large-scale, weakly labeled data to integrate a broad range of remote sensing visual knowledge into the LLM.
The second stage involves multi-task pretraining, where the vision encoder was further trained and the LLM is fine-tuned using LoRA \cite{hu2021lora} with multi-task instruction data, thereby enhancing multimodal and multi-task processing capabilities.
In the third stage, the LLM was fine-tuned with more complex instruction data to activate the model’s multi-task solving and reasoning capabilities.
H2RSVLM \cite{h2rsvlm} first underwent a pretraining phase, and during the supervised fine-tuning phase, the vision encoder was frozen while the LLM and projection layer were fine-tuned. 

\lightblue{Only Fine-tuning. }
Another part of the work involves obtaining remote sensing MLLMs using only instruction fine-tuning methods. This is further divided into single-stage fine-tuning and multi-stage fine-tuning.

To transfer models pretrained on natural scene domain to remote sensing domain, many studies have used a single-stage instruction fine-tuning training method. 
RS-LLaVA \cite{rs-llava} only employed LoRA to conduct instruction fine-tuning on the LLM. 
In addition to using LoRA for instruction fine-tuning on specific components of the large language model, some approaches \cite{geochat, skysensegpt, rsgpt} also unfreeze the image-text alignment layers, such as the Q-Former or MLP projector layer, further enhancing the alignment between image and text features. 
Due to the multimodal nature and complexity of remote sensing, single-stage instruction fine-tuning may not achieve optimal alignment. As a result, some studies have explored multi-stage fine-tuning methods.

SkyEyeGPT \cite{skyeyegpt} designed a two-stage instruction tuning approach, consisting of remote sensing image-text alignment and multi-task conversational fine-tuning to enhance instruction-following and multi-turn conversational abilities, respectively.
In the remote sensing image-text alignment stage, the model is trained using single-task image-text instructions, while in the multi-task conversational fine-tuning stage, multi-task conversational instructions are used to fine-tune the model.
EarthGPT \cite{earthgpt} is trained on natural scene data by unfreezing the self-attention and RMSNorm layers to achieve vision-language alignment and deep cross-modal understanding. 
In the second stage, inspired by LLaMA-Adapter V2 \cite{gao2023llama}, instruction fine-tuning in the remote sensing domain is performed by introducing learnable parameters with linear transformations in the linear layer to enhance the model's adaptability to remote sensing tasks. 
Popeye \cite{popeye} adopted distinct parameter optimization methods in the visual-language alignment and ship domain adaptation stages. 
In the visual-language alignment stage, the LoRA method is employed to fine-tune the LLM on general datasets.
In the second stage, to adapt to multi-source, multi-granularity ship detection data, the concept of LLaMA-Adapter V2 is integrated into LoRA fine-tuning and trained on the newly constructed MMShip dataset.

\section{Training Data}
\label{sec:training_data}

Training data plays a critical role in the development and success of foundation models. The performance of these models is inherently tied to the volume and diversity of the datasets on which they are trained. 
Prominent foundation models, such as CLIP \cite{clip}, DINOv2 \cite{dinov2}, SAM \cite{sam}, and the GPT series \cite{gpt, gpt2, gpt3}, have achieved remarkable results by leveraging extensive and heterogeneous datasets. 
CLIP \cite{clip}, for example, was trained on a newly curated dataset consisting of 400 million image-text pairs, sourced from a variety of publicly available resources across the internet. 
This large-scale dataset enabled CLIP to demonstrate exceptional performance across a wide range of multimodal tasks. 
Similarly, DINOv2 \cite{dinov2} was trained on a dataset of 142 million images, constructed by combining carefully curated datasets with publicly available images scraped from the web. 
SAM \cite{sam} was trained using the SA-1B dataset, which includes over 1 billion masks derived from 11 million licensed and privacy-preserving images. 
The dataset is self-supervisedly annotated, enabling SAM to learn highly generalized segmentation capabilities.
The GPT series \cite{gpt, gpt2, gpt3}, trained on large-scale text datasets curated from diverse publicly available online sources, demonstrate impressive generalization across a wide array of applications.

Similarly, training data is crucial for remote sensing foundation models, as it directly affects their performance and generalization capabilities.
Unlike traditional vision or multimodal models, which predominantly rely on text and RGB images, remote sensing data comprises a wide array of modalities, including optical, hyperspectral, and SAR imagery.
This diversity introduces significant complexity to both data processing and model training.

In the subsequent sections, we will review the training datasets utilized by various categories of remote sensing foundation models. 
Section \ref{sec:vfm} will examine the datasets employed by vision foundation models, with Section \ref{sec:collection} addressing custom-collected datasets and Section \ref{sec:publicly} focusing on publicly available datasets. 
Section \ref{sec:mmfm} will explore the datasets used by multimodal foundation models, with Section \ref{sec:vlm} dedicated to VLMs and Section \ref{sec:mllm} focusing on MLLMs. 

\subsection{Training Datasets for Vision Foundation Models}
\label{sec:vfm}

\begin{table*}[h!]
\centering
\caption{Summary of Training Datasets for Remote Sensing  Vision Foundation Model.}
\setlength\tabcolsep{0.4pt}%
\scriptsize
\begin{tabular*}{\linewidth}{@{}lccccccc@{}}
\toprule
Dataset & Year & Image & Size & \#Class & Resolution (m) & Modality & Models \\
\midrule

SARSim \cite{kusk2016synthetic} & 2016 & 21,168 & - & 7 & 0.1-0.3 & SAR & SAR-JEPA \cite{li2024predictinggradientbetterexploring} \\

fMoW \cite{fmow} & 2018 & $\sim$1,000,000 & - & 62 & 0.5 & RGB & \makecell{SatMAE \cite{satmae2022}, SatMAE++ \cite{noman2024rethinking},\\ Scale-MAE \cite{10377166}} \\

Sen12MS \cite{sen12ms} & 2019 & 180,662 & $256 \times 256$ & - & - & SAR, Multispectral & \makecell{RS-BYOL \cite{rs_byol_2022}, IaI-SimCLR \cite{Prexl_2023_CVPR},\\ msGFM \cite{han2024bridging}} \\

BigEarthNet-S2 \cite{sumbul2019bigearthnet} & 2019 & 590,326 & $20 \times 20 - 120 \times 120$ & 19 & 10-60 & Multispectral & SpectralGPT, S2MAE \cite{Li_2024_CVPR} \\

SAR-Ship \cite{wang2019sar} & 2019 & 39,729 & $256 \times 256$ & 1 & 3-25 & SAR & SAR-JEPA \cite{li2024predictinggradientbetterexploring} \\

SAMPLE \cite{lewis2019sar} & 2019 & 5,380 & - & 10 & 0.3 & SAR & SAR-JEPA \cite{li2024predictinggradientbetterexploring} \\

Seco \cite{Manas_2021_ICCV} & 2021 & $\sim$1,000,000 & - & - & 10-60 & Multispectral & Seco \cite{Manas_2021_ICCV} \\

MATTER \cite{Akiva_2022_CVPR} & 2021 & 14,857 & $1096 \times 1096$ & - & 10 & - & MATTER \cite{Akiva_2022_CVPR} \\

Levir-KR \cite{geokr_2022} & 2021 & 1,431,950 & $256 \times 256$ & 8 & 0.8-16 & RGB & GeoKR \cite{geokr_2022} \\

Million-AID \cite{millionaid} & 2021 & $\sim$1,000,000 & $110 \times 110 - 31,672 \times 31,672$ & 51 & 0.5-153 & RGB & \makecell{GeRSP \cite{10400411}, CMID \cite{10105625},\\ LeMeViT \cite{jiang2024lemevit}, msGFM \cite{han2024bridging}} \\

BigEarthNet-MM \cite{bigearthnetmm} & 2021 & 590,326 & $20 \times 20 - 120 \times 120$ & - & 10-60 & SAR, Multispectral & DINO-MM \cite{wang2022selfsupervised}\\

RingMo \cite{ringmo_2023} & 2022 & 2,096,640 & $448 \times 448$ & - & 0.3-30 &- & RingMo \cite{ringmo_2023} \\

TOV-RS \cite{10110958} & 2022 & 3,000,000 & - & 31 & - & RGB & TOV \cite{10110958} \\

SSL4EO-S12\cite{ssl4eo} & 2022 & $\sim$1,000,000 & $264 \times 264$ & - & 10 & SAR, Multispectral & \makecell{CROMA \cite{fuller2024croma}, FG-MAE \cite{wang2023feature},\\ DeCUR \cite{wang2023decurdecouplingcommon}} \\

SatlasPretrain \cite{satlaspretrain} & 2022 & 3,615,184 & $512 \times 512 - 8,192 \times 8,192$ & 137 & - & RGB, Multispectral & USatMAE \cite{irvin2023usat} \\

fMoW-S2 \cite{satmae2022} & 2022 & 712,874 & - & - & 10-60 & Multispectral & \makecell{SatMAE \cite{satmae2022}, SpectralGPT \cite{10490262}, \\ S2MAE \cite{Li_2024_CVPR}, SatMAE++ \cite{noman2024rethinking},\\ Scale-MAE \cite{10377166}} \\

MSAR \cite{chen2022large} & 2022 & 28,499 & $256 \times 256$ & 4 & 1 & SAR & SAR-JEPA \cite{li2024predictinggradientbetterexploring} \\

SkySense \cite{guo2024skysense} & 2023 & $\sim$21,500,000 & - & - & - & RGB, SAR, Multispectral & SkySense \cite{guo2024skysense}\\

GeoPile\cite{Mendieta_2023_ICCV} & 2023 & $\sim$600,000 & - & - & - & RGB & GFM \cite{Mendieta_2023_ICCV} \\

U-BARN \cite{10414422} & 2024 & $\sim$27,000 & $64 \times 64$ & - & 10 & Multispectral & U-BARN \cite{10414422} \\

GeoSense \cite{10378718} & 2024 & 8,916,233 & $224 \times 224$ & - & 0.05-150 & - & SMLFR \cite{10378718} \\

MMEarth \cite{nedungadi2024mmearth} & 2024 & $\sim$1,200,000 & $128 \times 128$ & - & 10 & RGB, SAR, Multispectral & MP-MAE \cite{nedungadi2024mmearth} \\

SAMRS \cite{wang2024samrs} & 2024 & 105,090 & $600 \times 600 - 1,024 \times 1,024$ & - & - & RGB & MTP \cite{wang2024mtp} \\

\bottomrule
\end{tabular*}
\label{tab:vfmdatasets}
\end{table*}

Vision foundation models primarily utilize image modality data for training, which can be sourced from both custom-collected and publicly available datasets. 
Some models are trained on custom datasets specifically gathered for particular applications or regions, enabling more targeted training that caters to the unique characteristics and requirements of the target environment. 
In contrast, other models rely on publicly available datasets, which provide standardized benchmarks and facilitate comparison across different methods. 
These datasets vary significantly in size, category, resolution, and modality, reflecting the diverse range of real-world scenarios encountered in remote sensing tasks.
For example, some datasets focus on high-resolution imagery, which is crucial for tasks requiring detailed spatial information, while others incorporate multispectral data to enable the modeling of diverse environmental conditions. 
For example, some datasets focus on high-resolution imagery, which is essential for tasks requiring detailed spatial information, while some others incorporate multispectral data to capture diverse semantic information.
The specific attributes of these datasets are summarized in Table \ref{tab:vfmdatasets}, demonstrating the variety and breadth of the data to train vision foundation models.

\subsubsection{Custom-collected Datasets}
\label{sec:collection}

Custom-collected datasets offer considerable advantages in training vision foundation models, as they can be meticulously designed to capture distinctive characteristics, regional features, or specific environmental conditions. This targeted approach enables models to better address the unique requirements of specialized tasks, thereby enhancing their adaptability and robustness.

\lightblue{Single-source Datasets.} 
SeCo \cite{Manas_2021_ICCV}, U-BARN \cite{10414422}, and MATTER \cite{Akiva_2022_CVPR} primarily utilized Sentinel-2 imagery for training. 
SeCo \cite{Manas_2021_ICCV} collected 1 million multi-spectral image patches gathered from 200,000 global locations, covering 12 spectral bands at varying resolutions (10 m, 20 m, and 60 m) across different time points to capture seasonal variations. 
U-BARN \cite{10414422} pretrained models on Sentinel-2 tiles from 13 regions in France, with edge, saturation, and cloud masks applied to improve image quality. 
MATTER \cite{Akiva_2022_CVPR} collected orthorectified Sentinel-2 images sourced from AWS, spanning diverse climates and regions over a three-year period, with cloud filtering applied to enhance data reliability. 
CtxMIM’s \cite{zhang2024ctxmimcontextenhancedmaskedimage} dataset was derived from WorldView-3 imagery collected via Google Earth Engine, capturing a range of Asian landscapes, such as cities, rivers, villages, and forests.

\lightblue{Multi-source Datasets.} SkySense \cite{guo2024skysense}, 
SMLFR \cite{10378718}, MP-MAE \cite{nedungadi2024mmearth}, RingMo \cite{ringmo_2023}, GeoKR \cite{geokr_2022}, and TOV \cite{10110958} integrated data from various remote sensing sources. 
SkySense \cite{guo2024skysense} integrated WorldView-3/4 optical imagery, Sentinel-2 multispectral time-series, and Sentinel-1 SAR data, providing comprehensive temporal and spectral perspectives.
The GeoSense dataset used in SMLFR \cite{10378718} includes imagery from multiple satellites (Sentinel-2, Gaofen, Landsat, and QuickBird) with resolutions ranging from 0.05 m to 150 m, making it suitable for a wide range of remote sensing applications.
MP-MAE \cite{nedungadi2024mmearth} proposed the MMEarth dataset, which consists of 12 geo-aligned modalities, including six pixel-level and six image-level modalities, gathered globally through Google Earth Engine.
The RingMo \cite{ringmo_2023} dataset is a large-scale collection used for self-supervised learning in remote sensing, featuring 2,096,640 images sourced from public datasets and China’s Gaofen-2 satellite.
These images span multi-source, multi-temporal, and multi-instance characteristics, with images cropped to 448$\times$448 pixels and resolutions varying from 0.3 to 30 meters.
GeoKR \cite{geokr_2022} utilized the Levir-KR dataset, containing over 1.4 million high-resolution satellite images from Gaofen satellites, covering various terrains, and converted to RGB format for training purposes.
TOV \cite{10110958} utilized two datasets: TOV-NI, which includes 1 million web-crawled natural images, and TOV-RS, a remote sensing dataset with both imbalanced and balanced versions, totaling up to 3 million samples.

\subsubsection{Publicly Available Datasets}
\label{sec:publicly}

Publicly available datasets offer standardized benchmarks, promoting consistent model comparison and evaluation across various approaches. 
Their diverse data sources enhance the generalization capability of models, ensuring robust performance across different contexts.

\lightblue{RGB Datasets.} Several models rely heavily on large-scale aerial RGB imagery for broad scene understanding and general-purpose feature extraction. 
For instance, models such as GeRSP \cite{10400411}, CMID \cite{10105625} and LeMeViT \cite{jiang2024lemevit} utilized the Million-AID dataset \cite{millionaid}, which contains over 1 million images across 51 categories, capturing diverse landscapes from Google Earth. 
Furthermore, Wang et al. \cite{wang2024mtp} leveraged the SAMRS dataset \cite{wang2024samrs}, which combines bounding box annotations from DOTA-V2 \cite{dota}, DIOR \cite{dior}, and FAIR1M \cite{fair1m}, producing a total of 105,090 images with 1,668,241 instances, all designed for segmentation tasks. 
Mendieta et al. \cite{Mendieta_2023_ICCV} introduced the GeoPile dataset, a diverse collection of approximately 600,000 labeled and unlabeled remote sensing images designed to improve geospatial foundation models. 
The dataset spans various GSD and includes data from different sources. This diversity ensures rich feature representation, enhancing model generalization across multiple geospatial tasks.

\lightblue{Multimodal Datasets.} 
Multimodal datasets are essential for training multimodal visual models like RS-BYOL \cite{rs_byol_2022}, IaI-SimCLR \cite{Prexl_2023_CVPR}, DINO-MM \cite{wang2022selfsupervised}, CROMA \cite{fuller2024croma}, DeCUR \cite{wang2023decurdecouplingcommon}, USatMAE \cite{irvin2023usat}, and FG-MAE \cite{wang2023feature}. 
These datasets facilitate comprehensive cross-modal feature learning by providing paired or triplet data from heterogeneous sensors. 

RS-BYOL and IaI-SimCLR utilized the Sen12MS dataset \cite{sen12ms}, which comprises 180,662 triplets of SAR, Sentinel-2 optical imagery, and MODIS land cover maps.
DINO-MM \cite{wang2022selfsupervised} proposed BigEarthNet-MM \cite{bigearthnetmm}, a multimodal extension of the BigEarthNet dataset linking Sentinel-1 and Sentinel-2 images.
CROMA \cite{fuller2024croma} and FG-MAE \cite{wang2023feature} employed the SSL4EOS12 dataset \cite{ssl4eo}, providing 1 million paired samples of Sentinel-1 and Sentinel-2 across various seasons.
DeCUR \cite{wang2023decurdecouplingcommon} integrated diverse multimodal datasets, including SSL4EOS12, GeoNRW \cite{baiergeonrw} for RGB-DEM fusion, and SUN-RGBD \cite{song2015sun} for RGB-depth analysis.
GeoNRW \cite{baiergeonrw} supplied 111,000 cropped RGB and digital elevation model (DEM) patches from North Rhine-Westphalia, Germany, enhancing elevation-based semantic interpretation, while SUN-RGBD \cite{song2015sun} offers 10,335 RGB-depth image pairs, augmenting depth-based scene recognition.

USatMAE \cite{irvin2023usat} was trained on the Satlas \cite{satlaspretrain} dataset, incorporating NAIP and Sentinel-2 imagery and pairing them based on spatial overlap and minimal temporal disparity.
Unlike these paired approaches, RSPrompter \cite{chen2023rsprompter} utilized distinct datasets, including WHU Building Extraction \cite{ji2018fully}, NWPU VHR-10 \cite{cheng2014multi}, and SSDD \cite{zhang2021sar}, each contributing unique modalities and categories.
MsGFM \cite{han2024bridging} proposed GeoPile-2, a comprehensive dataset combining RGB, SAR, and elevation data from Million-AID \cite{millionaid}, GeoPile \cite{Mendieta_2023_ICCV}, Sen12MS \cite{sen12ms}, and MDAS \cite{mdas}.
This multimodal integration enables general-purpose pretraining, significantly enhancing model performance across a variety of remote sensing tasks, including scene classification, object detection, and semantic segmentation.
Additionally, SatMAE \cite{satmae2022}, SatMAE++ \cite{noman2024rethinking}, and Scale-MAE \cite{10377166} were trained using two datasets: the fMoW \cite{fmow} RGB dataset and fMoW-S2 \cite{satmae2022}, an augmented dataset derived from fMoW that incorporates all 13 Sentinel-2 bands in addition to supplementary temporal data. 

\lightblue{Multispectral Datasets.} Models such as SpectralGPT \cite{10490262} and S2MAE \cite{Li_2024_CVPR} are specifically designed to exploit spectral and multispectral data for capturing fine-grained variations in land cover. 
They utilized fMoW-S2 \cite{satmae2022} and BigEarthNet-S2 \cite{sumbul2019bigearthnet}, exploiting 12 spectral bands from Sentinel-2 for in-depth spectral analysis. 

\lightblue{SAR-Based Datasets.} Models such as SAR-JEPA \cite{li2024predictinggradientbetterexploring} are predicated on the utilization of SAR-specific datasets to bolster feature extraction capabilities in radar imagery analysis. 
SAR-JEPA incorporated four principal SAR datasets: MSAR \cite{chen2022large}, SAR-Ship \cite{wang2019sar}, SARSim \cite{ kusk2016synthetic}, and SAMPLE \cite{lewis2019sar}. 
These datasets encompass a wide array of targets, notably including aircraft, ships, and military vehicles, which are imaged under a variety of conditions and resolutions.

\subsection{Training Datasets for Multimodal Foundation Models}
\label{sec:mmfm}

Multimodal foundational models typically leverage datasets that integrate image and text modalities for training, facilitating the learning of complex cross-modal relationships.
This training process primarily utilizes publicly available image-text paired datasets, which combine visual and textual components. 
In addition, Transformer-based MLLMs utilize various downstream visual task datasets to enhance their generalization capabilities, such as scene classification, object detection, segmentation, image captioning, and Visual Question Answering (VQA). 
The datasets are preprocessed into formats compatible with model requirements to ensure both compatibility and efficiency during training.
Nowadays, synthetic data has become a crucial component in the data flywheel of large models \cite{wang2024survey} and is widely used in the training of remote sensing multimodal foundational models. 
The specific information about these datasets is shown in Table \ref{tab:mfmdatasets}, while Table \ref{tab:mmmodeldata} details the models that utilize these datasets.
The subsequent sections will offer an in-depth introduction to the training datasets for VLMs and MLLMs. 

\begin{table*}[h!]
\centering
\caption{\centering Summary of Training Datasets for Remote Sensing Multimodal Foundation Models.}
\small
\setlength\tabcolsep{18pt}
\begin{tabular*}{\linewidth}{@{}lcccc@{}}
\toprule
Datasets         & Year & Image    & Resolution              & Models                                                               \\ 
\midrule

UCM \cite{yang2010bag}             & 2010 & 2,100       & 256 $\times$ 256        & RS-CLIP \cite{rs-clip}, EarthGPT \cite{earthgpt} \\ 

UCM-Caption \cite{qu2016deep}     & 2016 & 2,100       & 256 $\times$ 256        & \makecell{RemoteCLIP \cite{remoteclip}, S-CLIP \cite{s-clip},\\ SkyEyeGPT \cite{skyeyegpt}, EarthGPT \cite{earthgpt},\\ RS-LLaVA \cite{rs-llava}} \\ 

Sydney-Caption \cite{qu2016deep}  & 2016 & 613         & 500 $\times$ 500        & \makecell{S-CLIP \cite{s-clip}, SkyEyeGPT \cite{skyeyegpt},\\ EarthGPT \cite{earthgpt}} \\ 

NWPU-RESISC45 \cite{cheng2017remote}   & 2016 & 31,500      & 256 $\times$ 256        & \makecell{RS-CLIP \cite{rs-clip}, GeoChat \cite{geochat},\\ EarthGPT \cite{earthgpt}} \\ 

RSICD \cite{lu2017exploring}           & 2017    & 10,921      & 224 $\times$ 224        & \makecell{RemoteCLIP \cite{remoteclip}, S-CLIP \cite{s-clip},\\ SkyEyeGPT \cite{skyeyegpt}, EarthGPT \cite{earthgpt}} \\ 

DOTA \cite{dota}            & 2018 & 2,806       & -                       & \makecell{RemoteCLIP \cite{remoteclip}, GeoChat \cite{geochat},\\ EarthGPT \cite{earthgpt}} \\ 

fMoW \cite{fmow}            & 2018 & $\sim$1,000,000     & \makecell{110 $\times$ 110 - \\31,672 $\times$ 31,672}     & GeoRSCLIP \cite{georsclip} \\

RSVQA-LR \cite{lobry2020rsvqa}        & 2020 & 722         & 256 $\times$ 256        & \makecell{SkyEyeGPT \cite{skyeyegpt}, GeoChat \cite{geochat}, \\EarthGPT \cite{earthgpt}, RS-LLaVA \cite{rs-llava}} \\ 

RSVQA-HR \cite{lobry2020rsvqa}        & 2020 & 10,659      & 512 $\times$ 512        & SkyEyeGPT \cite{skyeyegpt} \\ 

DIOR \cite{dior}            & 2020 & 23,463      & 800 $\times$ 800        & \makecell{RemoteCLIP \cite{remoteclip}, GeoChat \cite{geochat}, \\EarthGPT \cite{earthgpt}} \\ 

RSIVQA \cite{zheng2021mutual}        & 2021 & 37,264      & -                       & \makecell{SkyEyeGPT \cite{skyeyegpt}, EarthGPT \cite{earthgpt}, \\RS-LLaVA \cite{rs-llava}} \\ 

FloodNet \cite{rahnemoonfar2021floodnet}        & 2021 & 2,343       & 4,000 $\times$ 3,000    & GeoChat \cite{geochat}, EarthGPT \cite{earthgpt} \\ 

MillionAID \cite{millionaid}      & 2021 & $\sim$1,000,000   & -                       & GeoRSCLIP \cite{georsclip} \\ 

NWPU-Caption \cite{cheng2022nwpu}    & 2022 & 31,500      & 256 $\times$ 256        & SkyEyeGPT \cite{skyeyegpt}, EarthGPT \cite{earthgpt} \\ 

RSITMD \cite{yuan2022exploring}          & 2022 & 4,743       & 256 $\times$ 256        & \makecell{RemoteCLIP \cite{remoteclip}, SkyEyeGPT \cite{skyeyegpt},\\ EarthGPT \cite{earthgpt}} \\ 

FAIR1M \cite{fair1m}          & 2022 & 16,488     & -                       & GeoChat \cite{geochat}, EarthGPT \cite{earthgpt} \\ 

RSVG \cite{sun2022visual}            & 2022 & 4,329       & -                       & SkyEyeGPT \cite{skyeyegpt} \\ 

DIOR-RSVG \cite{zhan2023rsvg}      & 2023 & 17,402      & 800 $\times$ 800        & SkyEyeGPT \cite{skyeyegpt}, EarthGPT \cite{earthgpt} \\ 

SkyScript \cite{skyscript}       & 2023 & $\sim$5,200,000   & -                       & SkyCLIP \cite{skyscript} \\ 

S2-100K \cite{satclip} & 2023 & $\sim$100,000 & 256 $\times$ 256 & SatCLIP \cite{satclip} \\

RS5M \cite{georsclip}            & 2023 & $\sim$5,000,000   & -                       & GeoRSCLIP \cite{georsclip} \\ 

RSICap \cite{rsgpt}          & 2023 & 2,585       & 512 $\times$ 512        & RSGPT \cite{rsgpt} \\ 

LHRS-Align \cite{lhrs-bot}      & 2024 & $\sim$1,150,000   & -                       & LHRS-Bot \cite{lhrs-bot} \\ 

LHRS-Instruct \cite{lhrs-bot}   & 2024 & $\sim$398,000     & -                       & LHRS-Bot \cite{lhrs-bot} \\ 

HqDC-1.4M \cite{h2rsvlm} & 2024 & $\sim$1,400,000  & - & H2RSVLM \cite{h2rsvlm} \\

HqDC-Instruct \cite{h2rsvlm} & 2024 & $\sim$30,000  & - & H2RSVLM \cite{h2rsvlm} \\

\bottomrule
\end{tabular*}
\label{tab:mfmdatasets}
\end{table*}
\begin{table*}[h!]
\centering
\caption{Datasets Used by Different Models Across Tasks.}
\setlength\tabcolsep{1pt}
\scriptsize
\begin{tabular*}{\linewidth}{@{}lcccccc@{}}
\toprule
\textbf{Model} & \textbf{Image Caption} & \textbf{Detection/Segmentation} & \textbf{Image Classification} & \textbf{Visual Question Answer} & \textbf{Visual Ground} & \textbf{Other} \\
\midrule

RemoteCLIP \cite{remoteclip} & \makecell{RSICD,\\ RSITMD, \\ UCM-Captions} & \makecell{DOTA, DIOR, \\HRRSD,  RSOD,\\ LEVIR, HRSC, \\ VisDrone, AU-AIR, \\S-Drone,  CAPRK, \\Vaihingen, Potsdam, \\ iSAID, LoveDA}  & - & - & - & - \\

GeoRSCLIP \cite{georsclip} & - & - & \makecell{BigEarthNet, \\ fMoW, MillionAID} & -
& - & \makecell{LAION2B-en,\\ LAION400M, \\ LAIONCOCO,\\ COYO700M, \\ CC3M, CC12M, \\ YFCC15M,  WIT, \\ Redcaps, SBU, \\ Visual Genome} \\

SkyCLIP \cite{skyscript} & - & - & - & - & - & \makecell{SkyScript} \\

S-CLIP \cite{s-clip} & \makecell{RSICD,\\ UCM-Caption, \\ Sydney-Caption}  & - & - & - & - & -\\

RS-CLIP \cite{rs-clip} & - & - & \makecell{UCM, WHU-RS19, \\ NWPU-RESISC45, AID} & - & - & -\\

GeoCLIP \cite{geoclip} & - & - & - & - & - & MediaEval Placing \\Tasks 2016\\

SatCLIP \cite{satclip} & - & - & - & - & - & S2-100K \\

SkyEyeGPT \cite{skyeyegpt}  & \makecell{RSICD, RSITMD, \\UCM-Captions, \\ Sydney-Captions, \\ NWPU-Captions} & - & - & \makecell{ERA-VQA, RSIVQA, \\ RSVQA-LR, RSVQA-HR} & \makecell{RSVG,\\ DIOR-RSVG} & \makecell{CapERA, RSPG,\\ DOTA-Conversa,\\ DIOR-Conversa, \\UCM-Conversa,\\  Sydney-Conversa} \\

GeoChat \cite{geochat} & - & \makecell{DOTA, DIOR, \\ FAIR1M} &  \makecell{NWPU-RESISC-45} & \makecell{RSVQA-LR,  FloodNet} & - & -\\

RSGPT \cite{rsgpt} & - & - & - & - & - & \makecell{RSICap} \\

EarthGPT \cite{earthgpt} & \makecell{RSICD, RSITMD, \\ UCM-Captions, \\ Sydney-Captions,  \\ NWPU-Captions} & 
\makecell{ NWPUVHR10,\\  FAIR1M, HRRSD, \\UCAS-AOD,  RSOD,\\DOTA,VisDrone, \\SSDD, HRISD,\\ AIR-SARShip-2.0,\\ DIOR, HIT-UAV, \\ Sea-shipping,\\ Infrared-security, \\ Aerial-mancar, \\Double-light-vehicle,\\ Oceanic ship}  & \makecell{NWPU-RESISC45, \\ UCM, WHU-RS19, \\RSSCN7,  DSCR,\\EuroSAT, FGSCR-42} &
\makecell{FloodNet, RSVQA-LR, \\ RSIVQA, CRSVQA} & \makecell{DIOR-RSVG} &
\makecell{LAION-400M, \\ COCO Caption} \\

LHRS-Bot \cite{lhrs-bot} & \makecell{RSICD, RSITMD, \\ UCM-Captions, \\ NWPU-Captions} & -  & \makecell{UCM, RSITMD,\\ NWPU-RESISC-45,\\fMoW, METER-ML} & \makecell{RSVQA-LR, RSVQA-HR} & \makecell{RSVG,\\ DIOR-RSVG} & \makecell{LHRS-Align,\\ LHRS-Instruct} \\

H2RSVLM \cite{h2rsvlm} & - & \makecell{DOTA, FAIR1M, \\LoveDA, MSAR,\\ GID, FBP,\\ DeepGlobe,\\ CrowdAI} & \makecell{fMoW, RSITMD,\\ MillionAID, \\ NWPU-RESISC-45, \\METER-ML, UCM} & \makecell{RSVQA-LR} & \makecell{DIOR-RSVG} & \makecell{CVUSA, CVACT, \\BANDON, MtS-WH} \\

RS-LLaVA \cite{rs-llava} & 
\makecell{UCM-Caption, UAV} & - & - & \makecell{RSVQA-LR, RSIVQA} & - & -\\

\bottomrule
\end{tabular*}
\label{tab:mmmodeldata}
\end{table*}

\subsubsection{Training Datasets for VLMs}
\label{sec:vlm}

VLMs such as RemoteCLIP \cite{remoteclip}, SkyCLIP \cite{skyscript}, S-CLIP \cite{s-clip}, GeoRSCLIP \cite{georsclip}, and RS-CLIP \cite{rs-clip} demonstrate diverse and innovative approaches to utilizing image-text datasets in remote sensing applications.
RemoteCLIP \cite{remoteclip} constructed a comprehensive training dataset by incorporating three image-text retrieval datasets alongside 10 object detection and 4 segmentation datasets. 
A rule-based approach was employed to transform bounding box annotations and labels into descriptive text, enabling richer semantic association for both detection and segmentation data. 
SkyCLIP \cite{skyscript} utilized data sourced from the Google Earth Engine, targeting RGB bands in satellite and aerial imagery to enhance semantic richness. 
By integrating detailed OpenStreetMap (OSM) information, SkyCLIP employed a two-stage tag classification method to ascertain which OSM tags can be visually represented in remote sensing images, ensuring both global representation and semantic diversity. 

S-CLIP \cite{s-clip} combined multiple image-text pair datasets and generate pseudo-labels for semi-supervised learning. 
GeoRSCLIP \cite{georsclip} curated its RS5M dataset from two primary sources. 
Initially, it processed 11 publicly available image-text paired datasets, applying remote sensing-specific keywords for filtering and ensuring quality through deduplication using advanced VLM. 
Furthermore, GeoRSCLIP generated captions for large-scale datasets with class-level labels using VLM to enhance dataset richness. 
RS-CLIP \cite{rs-clip} utilized a quartet of remote sensing datasets, including UCM \cite{yang2010bag}, WHU-RS19 \cite{dai2010satellite}, NWPU-RESISC45 \cite{cheng2022nwpu}, and AID \cite{xia2017aid}.  
By generating pseudo-labels with CLIP, RS-CLIP refined labeling through multiple iterations, progressively improving the accuracy of training data. 

Collectively, these models exemplify the varied, sophisticated approaches used to handle and leverage multimodal datasets, capturing the complex relationships between visual and textual data in the context of remote sensing. 

In addition to image-text datasets, there also exist image datasets paired with geospatial metadata.
For instance, GeoCLIP \cite{geoclip} leveraged the MediaEval Placing Tasks 2016 (MP-16) \cite{larson2017benchmarking} dataset for training, comprising 4.72 million geotagged images sourced from Flickr.
This large-scale dataset provides images with corresponding GPS coordinates, enabling effective training for geo-localization tasks.
SatCLIP \cite{satclip} employed the S2-100K dataset for pretraining, which comprises 100,000 tiles of 256$\times$256 pixel, multi-spectral (12-channel) Sentinel-2 satellite imagery, each paired with its corresponding centroid location.
The dataset is designed to enhance multi-task applicability and geographic generalization, offering a broad range of location features derived from multi-spectral satellite imagery.

\subsubsection{Training Datasets for MLLMs}
\label{sec:mllm}

Recent advancements in MLLMs within remote sensing have involved the creation of comprehensive datasets and the development of sophisticated preprocessing methodologies.

SkyEyeGPT \cite{skyeyegpt} was trained using the SkyEye-968k dataset, which combines reorganized public data with a small portion of manually verified generated content. 
This dataset includes single-task public image-text instructions and multi-task dialogue instructions constructed by re-arranging data from various tasks. 
GeoChat \cite{geochat} utilized an automated pipeline for generating multimodal instruction fine-tuning data. 
This process involves extracting attributes like color and position from existing remote sensing datasets, inserting them into predefined templates, and using Vicuna \cite{vicuna} to generate multi-turn question-answer sequences. 
RSGPT \cite{rsgpt} assembled the RSICap dataset using the DOTA \cite{dota} dataset, distinguished by its detailed annotations from remote sensing experts encompassing scene descriptions and visual reasoning.

EarthGPT \cite{earthgpt} developed the MMRS-1M dataset by integrating 34 publicly accessible remote sensing datasets, designed with semi-structured templates for forming question-answer pairs. 
Initially trained on general domain datasets like LAION-400M \cite{laion400m} and COCO Caption \cite{cococap}, EarthGPT was fine-tuned on MMRS-1M to specialize in multisensor visual understanding within remote sensing. 
Similarly, LHRS-Bot \cite{lhrs-bot} crafted datasets including LHRS-Align and LHRSInstruct, with image data sourced from GEE and geographic features from OpenStreetMap. 
Employing Vicuna-v1.5-13B, captions were generated, while a subset of 15,000 images was used with GPT-4 to create complex instruction data.
H2RSVLM \cite{h2rsvlm} developed five datasets, drawing images from sources like Million-AID \cite{millionaid} and DOTA-v2 \cite{dota}, including an image-text pair dataset and three instruction fine-tuning datasets. 
RS-LLaVA \cite{rs-llava} compiled its instruction dataset using UCM-caption \cite{qu2016deep}, UAV \cite{uav}, RSVQA-LR \cite{lobry2020rsvqa}, and RSIVQA \cite{zheng2021mutual}, integrating caption and VQA data to form dialogue-based instruction-answer formats. 

Collectively, these projects represent a robust effort to refine and expand the capabilities of MLLMs in the context of remote sensing, highlighting the innovative integration of diverse datasets and detailed annotation methodologies.

\section{Evaluation}
\label{sec:evaluation}

The deployment of vision and multimodal foundation models in remote sensing has seen considerable expansion, owing to their exceptional performance across a wide array of tasks.
Vision foundation models, typically pretrained on large-scale datasets, are frequently adapted for remote sensing applications through fine-tuning on task-specific datasets.
This fine-tuning process customizes the models to meet the particular demands of tasks such as land cover classification, object detection, and segmentation.
In contrast, multimodal foundation models, which integrate data from multiple modalities, such as text and imagery, can be directly applied to various tasks. 
By leveraging their cross-modal learning capabilities, these models can tackle complex tasks without necessitating task-specific fine-tuning. 

In the following sections, we offer a comprehensive overview of downstream tasks pertinent to both visual (in Section \ref{sec:eval_visual}) and multimodal (in Section \ref{sec:eval_multi}) foundation models. 
This overview encompasses task definitions, commonly employed datasets for evaluation, typical evaluation metrics, and the performance outcomes of various models. 
By examining these facets, our objective is to illuminate how different models perform in specific remote sensing applications, thereby providing insights into their strengths and limitations across diverse benchmarks.

\subsection{Vison Foundation Model}
\label{sec:eval_visual}
\subsubsection{Scene Classification}
\label{sec: task_vis_cls}
Scene classification in remote sensing is a crucial task that involves categorizing an entire image into one of several predefined scene types, such as forest, urban, agricultural, or water. 
This task is centered around identifying and analyzing global features within the image to ascertain its overall category. By understanding scene types, researchers can better interpret spatial distributions and patterns, contributing to a wide range of applications in environmental monitoring, urban planning, and resource management.

For evaluating scene classification tasks, several datasets are commonly utilized, including EuroSAT \cite{helber2019eurosat}, NWPU-RESISC45 \cite{cheng2017remote}, and AID \cite{xia2017aid}. Each dataset provides a distinct set of images that cater to varied scene types and resolutions. These datasets are described in Table \ref{tab:vfmsceneclassdatasets}. The primary evaluation metric for this task is accuracy, defined as the proportion of correctly classified samples over the total samples, offering a straightforward measure of model performance across all categories. The performance results of standard models on these datasets are detailed in Table 
\ref{tab:vfmsceneclass}, illustrating the model's capabilities in accurately classifying different scene types.

\begin{table}[h!]
\centering
\setlength\tabcolsep{10pt}%
\scriptsize
\caption{\centering Scene Classification Datasets for Vision Foundation Models.}

\begin{tabular*}{\linewidth}{@{}lcccc@{}}
\toprule
Dataset & Year & Quantity & Class & Resolution  \\
\midrule
NWPU-RESISC45 \cite{cheng2017remote} & 2016 & 31,500 & 45 & 256 $\times$ 256  \\
AID \cite{xia2017aid} & 2017 & 10,000 & 30 & 600 $\times$ 600 \\
EuroSAT \cite{helber2019eurosat} & 2018 & 27,000 & 10 & 64 $\times$ 64  \\
\bottomrule
\end{tabular*}

\label{tab:vfmsceneclassdatasets}
\end{table}
\begin{table*}[h!]
\centering
\setlength\tabcolsep{17pt}%
\caption{\centering Scene Classification Performances (Accuracy, \%) of Vision Foundation Models on EuroSAT, AID, and RESISC-45 Datasets. TR Indicates the Percentage of Data Used for Testing.}
\scriptsize
\begin{tabular*}{\linewidth}{@{}lcccccc@{}}
\toprule
\multirow{2}{*}{Method} & \multirow{2}{*}{Backbone} & \multirow{2}{*}{EuroSAT} & AID & AID & RESISC-45 & RESISC-45 \\
    &    &    & (TR=20\%)   & (TR=50\%)    & (TR=10\%) & (TR=20\%) \\
\hline
SatMAE \cite{satmae2022}& ViT-L & 95.74 & 95.02 & 96.94 & 91.72 & 94.10 \\
SwiMDiff \cite{10453587}& ResNet-18  & 96.10 & - & - & - &   \text{-} \\
GASSL \cite{Ayush_2021_ICCV}& ResNet-50 & 96.38 & 93.55 & 95.92 & 90.86 &  93.06 \\
GeRSP \cite{10400411}& ResNet-50  & \text{-} & - & - & - &  92.74 \\
SeCo \cite{Manas_2021_ICCV}&  ResNet-50  & 97.34  & 93.47 & 95.99 & 89.64 &  92.91 \\
CACo \cite{Mall_2023_CVPR}&  ResNet-50  & 97.77 & 90.88 & 95.05 & 88.28 &  91.94 \\
TOV \cite{10110958} &  ResNet-50  & \text{-} & 95.16 & 97.09 & 90.97 &  93.79 \\
RingMo \cite{ringmo_2023} & Swin-B & \text{-} & 96.90 & 98.34 & 94.25 &  95.67 \\
CMID \cite{10105625}&  Swin-B  & \text{-} & 96.11 & 97.79 & 94.05 &  95.53 \\
GFM \cite{Mendieta_2023_ICCV} & Swin-B & - & 95.47 & 97.09 & 92.73 & 94.64 \\
CSPT \cite{rs14225675} &  ViT-L & \text{-} & 96.30 & - & - &  95.62 \\
Usat \cite{irvin2023usat} &  ViT-L  & 98.37 & - &  & - &  \text{-} \\
Scale-MAE \cite{10377166} & ViT-L & 98.59 & 96.44 & 97.58 & 92.63 &  95.04 \\
CtxMIM \cite{zhang2024ctxmimcontextenhancedmaskedimage} & Swin-B & 98.69 & - & - & - &  \text{-} \\
SatMAE++ \cite{noman2024rethinking}& ViT-L & 99.04 & - & - & - &  \text{-} \\
SpectralGPT \cite{10490262}& ViT-B & 99.21 & - & - & - &  \text{-} \\
SkySense \cite{guo2024skysense}& Swin-H & \text{-} & 97.68 & 98.60 & 94.85 &  96.32 \\
MTP \cite{wang2024mtp}& InternImage-XL & 99.24 & - & - & - &  96.27 \\
RVSA \cite{rvsa_2023} & ViTAE-B & - & 97.03 & 98.50 & 93.93 & 95.69  \\
\bottomrule
\end{tabular*}
\label{tab:vfmsceneclass}
\end{table*}

\subsubsection{Object Detection}
\label{sec: task_vis_object}

Object detection in remote sensing imagery is a critical task that involves the identification and classification of specific objects, such as buildings, vehicles, and vessels, within satellite or aerial imagery.
It encompasses both horizontal object detection, which uses horizontal bounding boxes, and rotated object detection, where oriented bounding boxes are employed to effectively capture objects positioned at arbitrary angles. 
The ability to accurately detect and classify objects in diverse orientations and scales is essential for effective analysis and decision-making in remote sensing.

To evaluate object detection models, several datasets are frequently utilized, including Xview \cite{xview}, DIOR \cite{dior}, DIOR-R \cite{diorr}, FAIR1M \cite{fair1m}, and DOTA-v1.0 \cite{dota}, each providing unique scenes and object varieties. Detailed specifications of these datasets are further elaborated in Table \ref{tab:oddatasets}. A prevalent evaluation metric for object detection is mean Average Precision (mAP), which calculates the mean of AP across multiple Intersection over Union (IoU) thresholds and classes. 
This metric offers a comprehensive assessment of both precision and recall, reflecting the model’s proficiency in accurately identifying and classifying objects. 
The results of common models tested on these datasets are presented in Table \ref{tab:vfmhbbod} and Table \ref{tab:vfmobbod}, showcasing their effectiveness in tackling various detection challenges.


\begin{table}[h!]
\centering
\setlength\tabcolsep{3pt}%
\caption{\centering Object Detection Datasets for Vision Foundation Models.}
\scriptsize
\begin{tabular*}{\linewidth}{@{}lccccc@{}}
\toprule
Dataset & Year & Image & Class & Bbox-Type & Resolution  \\
\midrule
Xview \cite{xview} & 2018 & 846 & 60 & hbb & beyond 2,000 $\times$ 2,000 \\
DOTA \cite{dota} & 2018 & 2,806 & 15 & obb & varies in size  \\
DIOR \cite{dior} & 2020 & 23,463 & 20 & hbb & 800 $\times$ 800 \\
DIOR-R \cite{diorr} & 2022 & 23,463 & 20 & hbb & 800 $\times$ 800\\
FAIR1M \cite{fair1m} & 2022 & 16,488 & 37 & obb & varies in size  \\
\bottomrule
\end{tabular*}
\label{tab:oddatasets}
\end{table}
\begin{table}[h!]
\centering
\setlength\tabcolsep{16pt}%
\caption{\centering Horizontal Object Detection Results (mAP, \%) of Vision Foundation Models on Xview and DIOR Datasets.}
\scriptsize
\begin{tabular*}{\linewidth}{@{}lccc@{}}
\toprule
Method & Backbone & Xview &  DIOR \\
\hline
GASSL \cite{Ayush_2021_ICCV} & ResNet-50 & 17.70 & 67.40 \\
SeCo \cite{Manas_2021_ICCV} & ResNet-50 & 17.20 & - \\
CACO \cite{Mall_2023_CVPR} & ResNet-50 & 17.20 &  66.91 \\
CtxMIM \cite{zhang2024ctxmimcontextenhancedmaskedimage} & Swin-B & 18.80 & -  \\
TOV \cite{10110958} & ResNet-50 & - & 70.16 \\
SatMAE \cite{satmae2022} & ViT-L & - & 70.89 \\
CSPT \cite{rs14225675} & ViT-L & - & 71.70 \\
GeRSP \cite{10400411} & ResNet-50 & - & 72.20 \\
GFM \cite{Mendieta_2023_ICCV} & Swin-B & - & 72.84 \\
Scale-MAE \cite{10377166} & ViT-L & - & 73.81 \\
CMID \cite{10105625} & Swin-B & - & 75.11 \\
RingMo \cite{ringmo_2023} & Swin-B & - & 75.90 \\
SkySense \cite{guo2024skysense} & Swin-H & - & 78.73 \\
MTP \cite{wang2024mtp} & InternImage-XL & 18.20 &  78.00 \\
RVSA \cite{rvsa_2023} & ViTAE-B & - & 73.22 \\

\bottomrule
\end{tabular*}
\label{tab:vfmhbbod}
\end{table}
\begin{table}[h!]
\centering
\setlength\tabcolsep{3pt}%
\caption{\centering Rotated Object Detection Results (mAP, \%) of Vision Foundation Models on DIOR-R, FAIR1M-2.0, and DOTA-V1.0 Datasets.}
\scriptsize
\begin{tabular*}{\linewidth}{@{}lcccc@{}}
\toprule
Method & Backbone & DIOR-R & FAIR1M-2.0  &  DOTA-V1.0 \\
\hline
CACo \cite{Mall_2023_CVPR} & ResNet-50 & 64.10  &  47.83 & -  \\
RingMo \cite{ringmo_2023} & Swin-B & -  & 46.21 & -  \\
GASSL \cite{Ayush_2021_ICCV} & ResNet-50 &  65.65 & 48.15 & -  \\
SatMAE \cite{satmae2022} & ViT-L & 65.66  & 46.55 & -  \\
TOV \cite{10110958} & ResNet-50 & 66.33 & 49.62 & -  \\
CMID \cite{10105625} & Swin-B & 66.37 & 50.58  & 77.36  \\
Scale-MAE \cite{10377166} & ViT-L & 66.47 & 48.31 & -  \\
GFM \cite{Mendieta_2023_ICCV} & Swin-B & 67.67 & 49.69 & -  \\
SMLFR \cite{10378718} & ConvNeXt-L & 72.33  & -  & 79.33   \\
SkySense \cite{guo2024skysense} & Swin-H & 74.27 & 54.57 & -  \\
MTP \cite{wang2024mtp} & InternImage-XL & 72.17 & 50.93  &  80.77   \\
RVSA \cite{rvsa_2023} & ViTAE-B & 70.67 & - & 81.01  \\
\bottomrule
\end{tabular*}
\label{tab:vfmobbod}
\end{table}

\subsubsection{Semantic Segmentation}
\label{sec: task_vis_segmen}

Semantic segmentation in remote sensing involves assigning a class label to each pixel, resulting in a comprehensive map that distinguishes various types of land cover, such as vegetation, water, roads, and buildings. 
This task necessitates a pixel-level understanding of the spatial layout within the images, allowing for detailed analysis and categorization of land features.

To assess model performance in semantic segmentation, several datasets are frequently utilized, including SpaceNetv1 \cite{spacenet}, LoveDA \cite{loveda}, iSAID \cite{isaid}, DynamicEarthNet-Pl \cite{dynamicearthnet}, and DynamicEarthNet-S2 \cite{dynamicearthnet}. 
These datasets offer diverse scenes and are discussed in detail in Table \ref{tab:ssdatasets}. 
The primary evaluation metric for semantic segmentation tasks is mean Intersection over Union (mIoU), which computes the average IoU across all classes. 
For each class, IoU is determined by dividing the intersection of predicted and ground truth areas by their union, providing a precise measure of prediction accuracy at the pixel level. 
The performance of standard models on these datasets is presented in Table \ref{tab:vfmseg}.



\begin{table}[h!]
\centering
\caption{\centering Semantic Segmentation Datasets for Vision Foundation Models.}
\setlength\tabcolsep{4.5pt}%
\scriptsize
\begin{tabular*}{\linewidth}{@{}lccccc@{}}
\toprule
Dataset & Year & Image & Class  & Resolution  \\
\midrule
SpaceNetv1 \cite{spacenet} & 2018 & 6,940 & 1 & varies in size \\
iSAID \cite{isaid} & 2019 & 2,806 & 15 & varies in size  \\
\makecell[l]{DynamicEarthNet-PlanetFusion\\(Dyna.-Pla.) \cite{dynamicearthnet}} & 2022 & 54,750 & 7 & 1,024 $\times$ 1,024  \\
LoveDA \cite{loveda} & 2021 & 5,987 & 7 & 1,024 $\times$ 1,024  \\
\makecell[l]{DynamicEarthNet-Sentinel2\\(Dyna.-S2) \cite{dynamicearthnet}}  & 2022 & 54,750 & 7 & varies in size\\

\bottomrule
\end{tabular*}
\label{tab:ssdatasets}
\end{table}
\begin{table*}[h!]
\centering
\caption{\centering Semantic Segmentation Results (mIoU, \%) of Vision Foundation Models on SpaceNetv1, LoveDA, iSAID, and DynamicEarthNet Datasets.}
\setlength\tabcolsep{15pt}%
\scriptsize
\begin{tabular*}{\linewidth}{@{}lcccccc@{}}
\toprule
Method &  Backbone & SpaceNetv1 & LoveDA & iSAID & Dyna. -pla(val/test) & Dyna. S2(val/test)\\
\hline
SeCo \cite{Manas_2021_ICCV} & ResNet-50 & 77.09  & 43.63 & 57.20 & - & 29.40/39.80 \\
GASSL \cite{Ayush_2021_ICCV}  & ResNet-50 &  78.51 & 48.76 & 65.95 & 34.00/40.80 & 28.10/41.00\\
SatMAE \cite{satmae2022}  & ViT-L & 78.07 & - & 62.97 & 32.80/39.90 & 30.10/38.70 \\
RingMo \cite{ringmo_2023} & Swin-B & - & - & 67.20 & - &  \\
CMID \cite{10105625} & Swin-B & - & - & 66.21 & 36.40/43.50 & - \\
CACo \cite{Mall_2023_CVPR}  & ResNet-50 &  77.94 & 48.89 & 64.32 & 35.40/42.70 & 30.20/42.50 \\
TOV \cite{10110958}  & ResNet-50 & - & 49.70 & 66.24 & 32.10/37.80 & - \\
GeRSP \cite{10400411}  & ResNet-50 & - & 50.56 & - & - & - \\
SMLFR \cite{10378718} & ConvNext-L & - & 53.03 & -  & - & - \\
CtxMIM \cite{zhang2024ctxmimcontextenhancedmaskedimage}  & Swin-B & 79.47  & - & - & - & - \\
GFM \cite{Mendieta_2023_ICCV} & Swin-B & - & - & 66.62 & 36.70/45.60 & - \\
Scale-MAE \cite{10377166} & ViT-L & 78.90 & - & 65.77 & 34.00/41.70 & - \\
SkySense \cite{guo2024skysense} & Swin-H & - & - & 70.91 & 39.70/46.50 & 33.10/46.20 \\
MTP \cite{wang2024mtp} & InternImage-XL & 79.16 & 54.17 & - & - & - \\
RVSA \cite{rvsa_2023} & ViTAE-B & - & 52.44 & 64.49 & 34.30/44.40 & -  \\
\bottomrule
\end{tabular*}
\label{tab:vfmseg}
\end{table*}

\subsubsection{Change Detection}
\label{sec: task_vis_change}

Change detection in remote sensing is a critical task that aims to identify differences between two images of the same region taken at different times. 
This task involves detecting areas of change and capturing both temporal and spatial variations within the imagery. 

For evaluating change detection models, commonly used datasets include OSCD \cite{daudt2018urban} and LEVIR \cite{chen2020spatial} dataset. These datasets offer diverse scenarios for assessing change detection capabilities, with detailed characteristics presented in Table \ref{tab:cddatasets}. 
The predominant evaluation metric for this task is the F1 score, which is the harmonic mean of precision and recall. It balances false positives and false negatives, providing a comprehensive measure of a model's performance, especially in contexts where precision and recall hold equal weight. 
The performance results of standard models on these datasets are shown in Table \ref{tab:vfmcd}, illustrating their effectiveness in accurately identifying change across varied landscapes.


\begin{table}[h!]
\centering
\caption{\centering Change Detection Datasets for Vision Foundation Models.}
\setlength\tabcolsep{18pt}%
\scriptsize
\begin{tabular*}{\linewidth}{@{}lccc@{}}
\toprule
Dataset & Year & ImagePairs & Resolution \\
\midrule
OSCD \cite{daudt2018urban} & 2018 & 24 & 600 $\times$ 600  \\
LEVIR \cite{chen2020spatial} & 2020 & 637  & 1,024 $\times$ 1,024  \\
\bottomrule
\end{tabular*}
\label{tab:cddatasets}
\end{table}
\begin{table}[h!]
\centering
\caption{\centering Change Detection Results (F1 Score, \%) of Vision Foundation Models on OSCD and LEVIR Datasets.}
\setlength\tabcolsep{14pt}%
\scriptsize
\begin{tabular*}{\linewidth}{@{}lccc@{}}
\toprule
Method & Backbone & OSCD  &  LEVIR  \\
\hline
GASSL \cite{Ayush_2021_ICCV} & ResNet-50 & 46.26  & -   \\
SeCo \cite{Manas_2021_ICCV}  & ResNet-50 & 47.67  & 90.14  \\
SwiMDiff \cite{10453587}  & ResNet-18 & 49.60 & -  \\
CACo \cite{Mall_2023_CVPR}  & ResNet-50 & 52.11  & -  \\
SatMAE \cite{satmae2022} & ViT-L & 52.76  & -  \\
CMID \cite{10105625} & Swin-B & - & 91.72 \\
RingMo \cite{ringmo_2023} & Swin-B & -  & 91.86  \\
SpectralGPT \cite{10490262} & ViT-B & 54.29  & -  \\
GFM \cite{Mendieta_2023_ICCV} & Swin-B & 59.82 & -  \\
Scale-MAE \cite{10377166} & - &  92.07 \\
SkySense \cite{guo2024skysense}  & Swin-H & 60.06  & 92.58  \\
MTP \cite{wang2024mtp} & InternImage-XL & 55.61  & 92.54  \\
RVSA \cite{rvsa_2023} & ViTAE-B & - & 90.86 \\
\bottomrule
\end{tabular*}
\label{tab:vfmcd}
\end{table}

\subsection{Multimodal Foundation Model}
\label{sec:eval_multi}

\subsubsection{Scene classification}
\label{sec: task_mllm_scene}

Scene classification in remote sensing, as performed in both multimodal and vision foundation models, involves categorizing images into predefined scene categories by analyzing their overall content and context. 
The tasks remain fundamentally similar across these model types. 
For multimodal models, additional datasets such as UCM \cite{yang2010bag}, WHU-RS19 \cite{dai2010satellite}, SIRI-WHU \cite{zhao2015dirichlet}, and PatternNet \cite{zhou2018patternnet} are utilized to broaden the scope of evaluation, allowing for more comprehensive analysis and comparison. Detailed information about these datasets is presented in Table \ref{tab:imgsceneclassdatasets}, and the corresponding performance metrics and rankings for multimodal models are provided in Table \ref{tab:imgsceneclass}.

\begin{table}[h!]
\centering
\caption{\centering Scene Classification Datasets for Multimodal Foundation Models.}
\setlength\tabcolsep{12pt}%
\scriptsize
\begin{tabular*}{\linewidth}{@{}lcccc@{}}
\toprule
Dataset & Year & Image & Class & Resolution  \\
\midrule
UCM \cite{yang2010bag} & 2010 & 2,100 & 21 & 256 $\times$ 256  \\
WHU-RS19 \cite{dai2010satellite} & 2012 & 1,005 & 19 & 600 $\times$ 600 \\
SIRI-WHU \cite{zhao2015dirichlet} & 2016 & 2,400 & 12 & 200 $\times$ 200  \\
PatternNet \cite{zhou2018patternnet} & 2018 & 30,400 & 38 & 256 $\times$ 256 \\

\bottomrule
\end{tabular*}
\label{tab:imgsceneclassdatasets}
\end{table}
\begin{table}[h!]
\centering
\setlength\tabcolsep{1pt}%
\caption{\centering Scene Classification Results (Accuracy, \%) of Multimodal Foundation Models on EuroSAT, NWPU-RESISC45, WHU-RS19, AID, and SIRI-WHU Datasets.}
\scriptsize
\begin{tabular*}{\linewidth}{@{}lcccccc@{}}
\toprule
Method & \makecell{EuroSAT \\ \cite{helber2019eurosat}} & \makecell{NWPU-RES\\ISC45 \cite{cheng2017remote}} & \makecell{WHU-RS19 \\ \cite{dai2010satellite}} & \makecell{ AID \\ \cite{xia2017aid}} & \makecell{SIRI-WHU \\\cite{zhao2015dirichlet}} \\
\midrule
EarthGPT \cite{earthgpt} & - & 93.84 & - & - & - \\
GeoChat \cite{geochat} & - & - & - & 72.03 & - \\
LHRS-Bot \cite{lhrs-bot} & 51.40 & 83.94 & 93.17 & 91.26 & 62.66 \\
$H^2RSVLM$ \cite{h2rsvlm} & - & 93.87 & 97.00 & 89.33 & 68.50 \\
RemoteCLIP \cite{remoteclip} & 59.94 & 79.84 & 94.66 & 87.90 & - \\
SkyCLIP-50 \cite{skyscript} & 51.33 & 70.94 & - & 71.70 & - \\
S-CLIP \cite{s-clip} & - & - & 86.30 & 70.80 & - \\
GeoRSCLIP \cite{georsclip} & 67.47 & 73.83 & - & 76.33 & - \\
RS-CLIP \cite{rs-clip} & - & 85.07 & 99.10 & 79.56 & - \\
\bottomrule
\end{tabular*}
\label{tab:imgsceneclass}
\end{table}

\subsubsection{Image Captioning}
\label{sec: task_mllm_caption}

Image captioning aims to automatically generate concise and accurate descriptive text for a given image. This task requires models to identify not only the main objects and activities within the image but also their relationships and the overall context, effectively translating these visual contents into natural language. In the field of remote sensing, image captioning involves describing elements such as natural landforms, meteorological conditions, and buildings from satellite or aerial imagery, producing precise and content-rich textual descriptions.

For evaluating image captioning in remote sensing, commonly used datasets include UCM-Caption \cite{qu2016deep}, Sydney-Caption \cite{qu2016deep}, RSICD \cite{lu2017exploring}, NWPU-Caption \cite{cheng2022nwpu}, and RSITMD \cite{yuan2022exploring}. Detailed specifications of these datasets are shown in Table \ref{tab:imgcapdata}. 
Common evaluation metrics for this task comprise:

BLEU: An automatic translation quality assessment metric, BLEU scores are based on the overlap of n-grams between generated text and reference texts, with BLEU-1 through BLEU-4 progressively examining longer word sequence matches, thereby capturing complex linguistic structures.

METEOR: A word-alignment-based evaluation metric that accounts for synonyms and sentence structure, enabling semantic matching beyond exact vocabulary.

ROUGE-L: Used to evaluate summary or translation quality, ROUGE-L focuses on the Longest Common Subsequence (LCS), assessing recall and precision, thereby emphasizing content completeness and fluency.

CIDEr: Specifically designed for image description tasks, CIDEr evaluates the quality of generated descriptions based on the consensus among human-generated descriptions.

The performance of various models on these evaluation datasets is presented in Table \ref{tab:img_cap}.


\begin{table}[h!]
\centering
\setlength\tabcolsep{8pt}%
\caption{\centering Image Captioning Datasets for Multimodal Foundation Models.}
\scriptsize
\begin{tabular*}{\linewidth}{@{}lcccc@{}}
\toprule
Dataset & Year & Image & Sentence & Resolution \\
\midrule
UCM-Captions \cite{qu2016deep} & 2016 & 2,100 & 10,500 & 256 $\times$ 256  \\
Sydney-Captions \cite{qu2016deep} & 2016 & 613 & 3,065 & 500 $\times$ 500\\
RSICD \cite{lu2017exploring} & 2017 & 10,921 & 54,605 & 224 $\times$ 224  \\
NWPU-Captions \cite{cheng2022nwpu} & 2022 & 31,500 & 157,500 & 256 $\times$ 256 \\
RSITMD \cite{yuan2022exploring} & 2022 & 4,743 & 23,715 & 256 $\times$ 256  \\
\bottomrule
\end{tabular*}
\label{tab:imgcapdata}
\end{table}
\begin{table*}[h!]
\centering
\setlength\tabcolsep{9.6pt}%
\caption{\centering Image Captioning Results (BLEU, METEOR, ROUGE\_L, CIDEr) of Multimodal Foundation Models on UCM-Captions, Sydney-Captions, RSICD, and NWPU-Caption Datasets.}
\scriptsize
\begin{tabular*}{\linewidth}{@{}llccccccc@{}}
\toprule
Dataset & Model & BLUE-1 & BLUE-2 & BLUE-3 & BLUE-4 & METEOR & ROUGE\_L & CIDEr \\
\midrule
\multirow{8}{*}{\makecell{UCM-Captions   \cite{qu2016deep}}} 
& SkyEyeGPT \cite{skyeyegpt} & 90.71 & 85.69 & 81.56 & 78.41 & 46.24 & 79.49 & 236.75 \\
& RSGPT \cite{rsgpt} & 86.12 & 79.14 & 72.31 & 65.74 & 42.21 & 78.34 & 333.23 \\
& RS-LLaVA7B \cite{rs-llava} & 88.70 & 82.88 & 77.70 & 72.84 & 47.98 & 85.17 & 349.43 \\
& RS-LLaVA13B \cite{rs-llava} & 90.00 & 84.88 & 80.30 & 76.03 & 49.21 & 85.78 & 355.61 \\
& RS-CapRet \cite{rs-cap} & 84.30 & 77.90 & 72.20 & 67.00 & 47.20 & 81.70 & 354.80 \\
& BITA \cite{yang2024bootstrapping} & 88.89 & 83.12 & 77.30 & 71.87 & 46.88 & 83.76 & 384.50 \\
& SCST \cite{zhang2023multi} & 87.27 & 80.96 & 75.51 & 70.39 & 46.52 & 82.58 & 371.29 \\
\midrule
\multirow{5}{*}{\makecell{Sydney-Captions  \cite{qu2016deep}}} 
& SkyEyeGPT \cite{skyeyegpt} & 91.85 & 85.64 & 80.88 & 77.40 & 46.62 & 77.74 & 181.06 \\
& RSGPT \cite{rsgpt} & 82.26 & 75.28 & 68.57 & 62.23 & 41.37 & 74.77 & 273.08 \\
& RS-CapRet \cite{rs-cap} & 78.70 & 70.00 & 62.80 & 56.40 & 38.80 & 70.70 & 239.20 \\
& SCST \cite{zhang2023multi} & 76.43 & 69.19 & 62.83 & 57.25 & 39.46 & 71.72 & 281.22 \\
\midrule
\multirow{6}{*}{RSICD \cite{lu2017exploring}} 
& SkyEyeGPT \cite{skyeyegpt} & 86.71 & 76.66 & 67.31 & 59.99 & 35.35 & 62.63 & 83.65 \\
& RSGPT \cite{rsgpt} & 70.32 & 54.23 & 44.02 & 36.83 & 30.10 & 53.34 & 102.94 \\
& RS-CapRet \cite{rs-cap} & 72.00 & 59.90 & 50.60 & 43.30 & 37.00 & 63.30 & 250.20 \\
& BITA \cite{yang2024bootstrapping} & 77.38 & 66.54 & 57.65 & 50.36 & 41.99 & 71.74 & 304.53 \\
& SCST \cite{zhang2023multi} & 78.36 & 66.79 & 57.74 & 50.42 & 36.72 & 67.30 & 284.36 \\
\midrule
\multirow{3}{*}{\makecell{NWPU-Caption   \cite{cheng2022nwpu}}} 
& EarthGPT \cite{earthgpt} & 87.10 & 78.70 & 71.60 & 65.50 & 44.50 & 78.20 & 192.60 \\
& RS-CapRet \cite{rs-cap} & 87.10 & 78.70 & 71.70 & 65.60 & 43.60 & 77.60 & 192.90 \\
& BITA \cite{yang2024bootstrapping} & 88.54 & 80.70 & 73.76 & 67.60 & 45.27 & 78.53 & 197.04 \\
\bottomrule
\end{tabular*}
\label{tab:img_cap}
\end{table*}

\subsubsection{Visual Question Answering}
\label{sec: task_mllm_vqa}

Visual Question Answering (VQA) in remote sensing involves generating answers to questions based on image content analysis. This requires models to identify and understand elements within an image and their relationships, integrating visual data with the context of the questions posed. In remote sensing, VQA tasks often address inquiries about features or events captured in satellite or aerial imagery, such as identifying land use types, assessing meteorological conditions, or counting specific objects.

Common datasets for evaluating VQA in remote sensing include RSVQA-LR \cite{lobry2020rsvqa}, RSVQA-HR \cite{lobry2020rsvqa}, FloodNet \cite{rahnemoonfar2021floodnet}, and RSIVQA \cite{zheng2021mutual}, as detailed in Table \ref{tab:vqadata}. The primary evaluation metric is accuracy, which measures the proportion of correct answers generated by the system, particularly useful for questions with definitive answers. The performances on these datasets are detailed in Table \ref{tab:lrvqa} and Table \ref{tab:hrvqa}.


\begin{table}[h!]
\centering
\setlength\tabcolsep{8pt}%
\caption{\centering Visual Question Answering Datasets for Multimodal Foundation Models.}
\scriptsize
\begin{tabular*}{\linewidth}{@{}lcccc@{}}
\toprule
Dataset & Year & Image & QA Pair & Resolution \\
\midrule
RSVQA-LR \cite{lobry2020rsvqa} & 2020 & 772 & 770,232 & 256 $\times$ 256 \\
RSVQA-HR \cite{lobry2020rsvqa} & 2020 & 10,659 & 1,066,316 & 512 $\times$ 512  \\
FloodNet-VQA \cite{rahnemoonfar2021floodnet} & 2021 & 1,448 & 4,511 & 4,000 $\times$ 3,000\\
RSIVQA \cite{zheng2021mutual} & 2021 & 37,264 & 111,693 & varies in size  \\
\bottomrule
\end{tabular*}
\label{tab:vqadata}
\end{table}

\begin{table}[h!]
\centering
\caption{\centering Visual Question Answering Results (Accuracy, \%) of Multimodal Foundation Models on the RSVQA-LR Test Set \cite{lobry2020rsvqa} Dataset.}
\setlength\tabcolsep{1pt}%
\scriptsize
\begin{tabular*}{\linewidth}{@{}lccccc@{}}
\toprule
Method & Count & Presence & Comparison & Rural/Urban & Avg. Accuracy \\
\midrule
SkyEyeGPT \cite{skyeyegpt} & - & 88.93 & 88.63 & 75.00 & 84.19 \\
RSGPT \cite{rsgpt} & - & 91.17 & 91.70 & 94.00 & 92.29 \\
GeoChat \cite{geochat} & - & 91.09 & 90.33 & 94.00 & 91.81 \\
LHRS-Bot \cite{lhrs-bot} & - & 88.51 & 90.00 & 89.07 & 89.19 \\
$H^2RSVLM$ \cite{h2rsvlm} & - & 89.58 & 89.79 & 88.00 & 89.12 \\
RS-LLaVA7B \cite{rs-llava} & 74.38 & 92.80 & 91.33 & 94.00 & 88.13 \\
RS-LLaVA13B \cite{rs-llava} & 73.76 & 92.27 & 91.37 & 95.00 & 88.10 \\
\bottomrule
\end{tabular*}
\label{tab:lrvqa}
\end{table}
\begin{table}[h!]
\centering
\caption{\centering Visual Question Answering Results (Accuracy, \%) of Multimodal Foundation Models on the RSVQA-HR Dataset.}
\setlength\tabcolsep{2pt}%
\scriptsize
\begin{tabular*}{\linewidth}{@{}llccc@{}}
\toprule
Dataset & Method & Presence & Comparison & Avg. Accuracy \\
\midrule
\multirow{2}{*}{\makecell{RSVQA-HR \\ Test Set 1 \cite{lobry2020rsvqa}}} 
& SkyEyeGPT \cite{skyeyegpt} & 84.95 & 85.63 & 85.29 \\
& RSGPT \cite{rsgpt} & 91.86 & 92.15 & 92.00 \\
\midrule
\multirow{6}{*}{\makecell{RSVQA-HR \\ Test Set 2 \cite{lobry2020rsvqa}}} 
& SkyEyeGPT \cite{skyeyegpt} & 83.50 & 80.28 & 81.89 \\
& RSGPT \cite{rsgpt} & 89.87 & 89.68 & 89.78 \\
& GeoChat \cite{geochat} & 58.45 & 83.19 & 70.82 \\
& EarthGPT \cite{earthgpt} & 62.77 & 79.53 & 71.15 \\
& LHRS-Bot \cite{lhrs-bot} & 92.57 & 92.53 & 92.55 \\
& $H^2RSVLM$ \cite{h2rsvlm} & 65.00 & 83.70 & 74.35 \\
\bottomrule
\end{tabular*}
\label{tab:hrvqa}
\end{table}

\subsubsection{Visual Grounding}
\label{sec: task_mllm_grounding}

Visual grounding is the task of locating reference objects within images based on natural language descriptions. 
This requires models to understand descriptive statements and analyze image content to accurately identify and mark objects specified in the descriptions. 
In remote sensing, visual grounding tasks often involve generating bounding boxes for specific objects, such as buildings and vehicles, within remote sensing images, guided by given descriptions.

Key datasets used for evaluating visual grounding in remote sensing include RSVG \cite{sun2022visual} and DIOR-RSVG \cite{zhan2023rsvg}, with details provided in Table \ref{tab:imggrouddata}. The primary evaluation metric is Accuracy@0.5, where a prediction is deemed accurate if the bounding box overlaps more than 0.5 Intersection over Union (IoU) with the true bounding box. The performances on these datasets is documented in Table \ref{tab:visualgrounding}.


\begin{table}[h!]
\centering
\setlength\tabcolsep{5.5pt}%
\caption{\centering Visual Grounding Datasets for Multimodal Foundation Models.}
\scriptsize
\begin{tabular*}{\linewidth}{@{}lcccc@{}}
\toprule
Dataset & Year & Image & Image Query Pair & Resolution  \\
\midrule
RSVG \cite{sun2022visual} & 2022 & 4,329 & 7,933 & varies in size \\
DIOR-RSVG \cite{zhan2023rsvg} & 2023 & 17,402 & 38,320 & 800 $\times$ 800  \\
\bottomrule
\end{tabular*}
\label{tab:imggrouddata}
\end{table}
\begin{table}[h!]
\centering
\setlength\tabcolsep{31pt}%
\caption{\centering Visual Grounding Results (Accuracy@0.5, \%) of Multimodal Foundation Models on RSVG and DIOR-RSVG Datasets.}
\scriptsize
\begin{tabular*}{\linewidth}{@{}llccccc@{}}
\toprule
Method & RSVG & DIOR-RSVG  \\
\midrule
EarthGPT \cite{earthgpt} & - & 76.65 \\
SkyEyeGPT \cite{skyeyegpt} & 70.50 & 88.59 \\
LHRS-Bot \cite{lhrs-bot} & 73.45 & 88.10 \\
$H^2RSVLM$ \cite{h2rsvlm} & - & 48.04 \\
\bottomrule
\end{tabular*}
\label{tab:visualgrounding}
\end{table}

\subsubsection{Cross-modal Retrieval}
\label{sec: task_mllm_retrieval}

Cross-modal retrieval involves retrieving data between images and texts, aiming to match a given image or text query with the most appropriate counterpart in the dataset. 
This task demands the model to effectively analyze and interpret data from two different modalities, facilitating accurate information retrieval. In remote sensing, cross-modal retrieval typically focuses on aligning remote sensing images with their descriptive texts.

For this task, commonly used datasets include RSICD \cite{lu2017exploring}, RSITMD \cite{yuan2022exploring}, and UCM-Captions \cite{qu2016deep}. The primary evaluation metric is Recall@K, which measures the proportion of relevant items retrieved within the top K results, compared to the total number of relevant items available. Performance outcomes for various models on these datasets are detailed in Table \ref{tab:cmretrival}.

\begin{table*}[tp]
\centering
\caption{\centering Cross-Modal Retrieval Results (Recall@K, \%) of Multimodal Foundation Models on RSICD, RSITMD, and UCM-Captions Datasets.}
\setlength\tabcolsep{10pt}%
\scriptsize
\begin{tabular*}{\linewidth}{@{}llccccccc@{}}
\toprule
Dataset & Method & I2T R@1 & I2T R@5 & I2T R@10 & T2I R@1 & T2I R@5 & T2I R@10 & Mean Recall \\
\midrule
\multirow{6}{*}{RSICD \cite{lu2017exploring}} & RemoteCLIP \cite{remoteclip} & 18.39 & 37.42 & 51.05 & 14.73 & 39.93 & 56.58 & 36.35 \\
& SkyCLIP-30 \cite{skyscript} & 8.97 & 24.15 & 37.97 & 5.85 & 20.53 & 33.53 & 21.84 \\
& S-CLIP \cite{s-clip} & 4.20 & 18.40 & - & 4.20 & 16.80 & - & - \\
& GeoRSCLIP \cite{georsclip} & 21.13 & 41.72 & 55.63 & 15.59 & 41.19 & 57.99 & 38.87 \\
& RS-CapRet \cite{rs-cap} & - & - & - & 10.25 & 31.62 & 48.53 & - \\
& PE-RSITR \cite{yuan2023parameter} & 14.13 & 31.51 & 44.78 & 11.63 & 33.92 & 50.73 & 31.12 \\
\midrule
\multirow{4}{*}{RSITMD \cite{yuan2022exploring}} & RemoteCLIP \cite{remoteclip} & 28.76 & 52.43 & 63.94 & 23.76 & 59.51 & 74.73 & 50.52 \\
& SkyCLIP-30 \cite{skyscript} & 11.73 & 33.19 & 47.35 & 10.19 & 32.47 & 49.08 & 30.67 \\
& GeoRSCLIP \cite{georsclip} & 32.30 & 53.32 & 67.92 & 25.04 & 57.88 & 74.38 & 51.81 \\
& PE-RSITR \cite{yuan2023parameter} & 23.67 & 44.07 & 60.36 & 20.10 & 50.63 & 67.97 & 44.47 \\
\midrule
\multirow{5}{*}{\makecell{UCM-Captions \\ 
 \cite{qu2016deep}}} & RemoteCLIP \cite{remoteclip} & 19.05 & 54.29 & 80.95 & 17.71 & 62.19 & 93.90 & 54.68 \\
& SkyCLIP-30 \cite{skyscript} & 38.57 & 84.29 & 93.81 & 31.83 & 64.19 & 81.96 & 65.78 \\
& S-CLIP \cite{s-clip} & 11.60 & 45.70 & - & 11.10 & 43.50 & - & - \\
& RS-CapRet \cite{rs-cap} & - & - & - & 16.10 & 56.29 & 90.76 & - \\
& PE-RSITR \cite{yuan2023parameter} & 22.71 & 55.81 & 80.33 & 18.82 & 62.84 & 93.72 & 55.71 \\
\bottomrule
\end{tabular*}
\label{tab:cmretrival}
\end{table*}

\section{Challenges}
\label{sec:Challenges}

\subsection{Long-Tail Distribution Data}

Despite the abundance of samples in remote sensing data, the long-tail problem \cite{zhang2023deep, liu2021self} within natural data distributions remains an inevitable challenge. This issue manifests prominently in the long-tail distribution of scenes and targets within the remote sensing domain, hindering the trained baseline model's ability to achieve satisfactory performance in relation to rare scenes and targets. Within natural image scenes, a variety of methods \cite{jiang2021self, zhou2022contrastive} have been utilized to tackle long-tail data issues. It is promising to anticipate the extension of these methodologies into the domain of remote sensing.

\subsection{Multimodal Integration}

The trend towards utilizing unified models \cite{zhu2022uni, wang2023one} is gaining prominence in current research. This involves refining techniques for the integration and processing of diverse data types, such as the fusion of optical and radar imagery, to derive deeper and more comprehensive insights. The OFA-Net \cite{xiong2024all}, which showcases the integration of multimodal data, emerges as a beacon of promise, guiding future models to emulate and further refine this approach. In this context, it is imperative to explore advanced self-supervised learning (SSL) methods that can adeptly leverage multimodal remote sensing data.

\subsection{Efficient Architecture}

Given that the resolution of remote sensing images significantly exceeds that of natural images, the training and inference overhead of remote sensing foundation models tends to be greater. Furthermore, in certain drone applications \cite{osco2021review}, an immediate response is imperative. These factors have driven the development of efficient remote sensing foundation models. Although some models have already been optimized for efficiency \cite{wang2024scaling, tseng2024lightweight}, there remains considerable potential for further improvement.

\subsection{Evaluation Benchmark}

Despite the emergence of numerous remote sensing foundation models which are designed to cater to a wide array of downstream remote sensing tasks, the methodologies for evaluating these models remain relatively constrained. Firstly, the current evaluation frameworks for remote sensing do not encompass all scenarios, including oceans \cite{kikaki2022marida} and polar \cite{gourmelon2022calving} regions. Secondly, tasks related to weather and climate \cite{sharifi2016assessment} evaluation are scarcely addressed in these evaluations. Lastly, there is a lack of unified evaluation benchmarks for remote sensing in the MLLM field, with most relying on traditional remote sensing benchmarks. This indicates that the existing remote sensing foundation models have not undergone comprehensive evaluation and analysis. Consequently, this has posed challenges for subsequent improvements and applications.

\clearpage

\bibliographystyle{IEEEtran}
\bibliography{paper}

\vfill

\end{document}